\documentclass[10pt,twocolumn,letterpaper]{article}

\usepackage[pagenumbers]{cvpr} 

\definecolor{cvprblue}{rgb}{0.21,0.49,0.74}
\usepackage[pagebackref,breaklinks,colorlinks,allcolors=cvprblue]{hyperref}

\usepackage{makecell}   
\usepackage{multirow}   
\usepackage[table]{xcolor}          

\def\paperID{N/A} 
\def\confName{CVPR}
\def\confYear{2026}

\definecolor{ForestGreen}{rgb}{0.13, 0.55, 0.13}
\definecolor{WildStrawberry}{rgb}{1.0, 0.26, 0.64}
\definecolor{ForestBlue}{rgb}{0.36, 0.54, 0.66}

\title{UTPTrack: Towards Simple and Unified Token Pruning for Visual Tracking}

\author{Hao Wu$^{1,2}$\thanks{Equal contribution.} \quad Xudong Wang$^{1}$\footnotemark[1] \quad Jialiang Zhang$^{1}$ \quad Junlong Tong$^{1,2,3}$ \\ Xinghao Chen$^{1,4}$ \quad Junyan Lin$^{1,4}$ \quad Yunpu Ma$^{5}$ \quad Xiaoyu Shen$^{1,2}$\thanks{Corresponding author.} \\
$^{1}$Institute of Digital Twin, Eastern Institute of Technology, Ningbo \\
$^{2}$Ningbo Key Laboratory of Spatial Intelligence and Digital Derivative \\
$^{3}$Shanghai Jiao Tong University \quad 
$^{4}$The Hong Kong Polytechnic University \\ 
$^{5}$Munich Center for Machine Learning, LMU Munich\\
{\tt\small haowu.ai.research@gmail.com \quad xyshen@eitech.edu.cn}
}

\begin{document}
\maketitle
\begin{abstract} 
One-stream Transformer-based trackers achieve advanced performance in visual object tracking but suffer from significant computational overhead that hinders real-time deployment. While token pruning offers a path to efficiency, existing methods are fragmented. They typically prune the search region, dynamic template, and static template in isolation, overlooking critical inter-component dependencies, which yields suboptimal pruning and degraded accuracy. To address this, we introduce \textbf{UTPTrack}, a simple and Unified Token Pruning framework that, for the first time, jointly compresses all three components. UTPTrack employs an attention-guided, token type-aware strategy to holistically model redundancy, a design that seamlessly supports unified tracking across multimodal and language-guided tasks within a single model. Extensive evaluations on 10 benchmarks demonstrate that UTPTrack achieves a new state-of-the-art in the accuracy-efficiency trade-off for pruning-based trackers, pruning 65.4\% of vision tokens in RGB-based tracking and 67.5\% in unified tracking while preserving 99.7\% and 100.5\% of baseline performance, respectively. This strong performance across both RGB and multimodal scenarios underlines its potential as a robust foundation for future research in efficient visual tracking. Code will be released at \href{https://github.com/EIT-NLP/UTPTrack}{https://github.com/EIT-NLP/UTPTrack}.
\end{abstract}    
\section{Introduction}
\label{sec:introduction}

Visual Object Tracking (VOT) aims to estimate the position and shape of a target throughout a video sequence given its initial state~\cite{chen2022visual, ye2022joint, chen2021transformer}. A typical tracker represents the target with a \emph{static template (ST)} from the first frame, optionally updates a \emph{dynamic template (DT)} to handle appearance variations, and localizes it within a cropped \emph{search region (SR)} in subsequent frames~\cite{yan2021learning, wang2023dynamic}. Owing to its dynamic nature, VOT requires strong spatio-temporal modeling~\cite{huang2019got, fan2019lasot, li2017tracking}. Recent advances in Transformer-based architectures have shown strong modeling capabilities and brought significant improvements in tracking performance~\cite{chen2021transformer, yan2021learning, 11395545, lin2022swintrack}.

\begin{figure}[t]
    \centering
    \includegraphics[width=\linewidth]{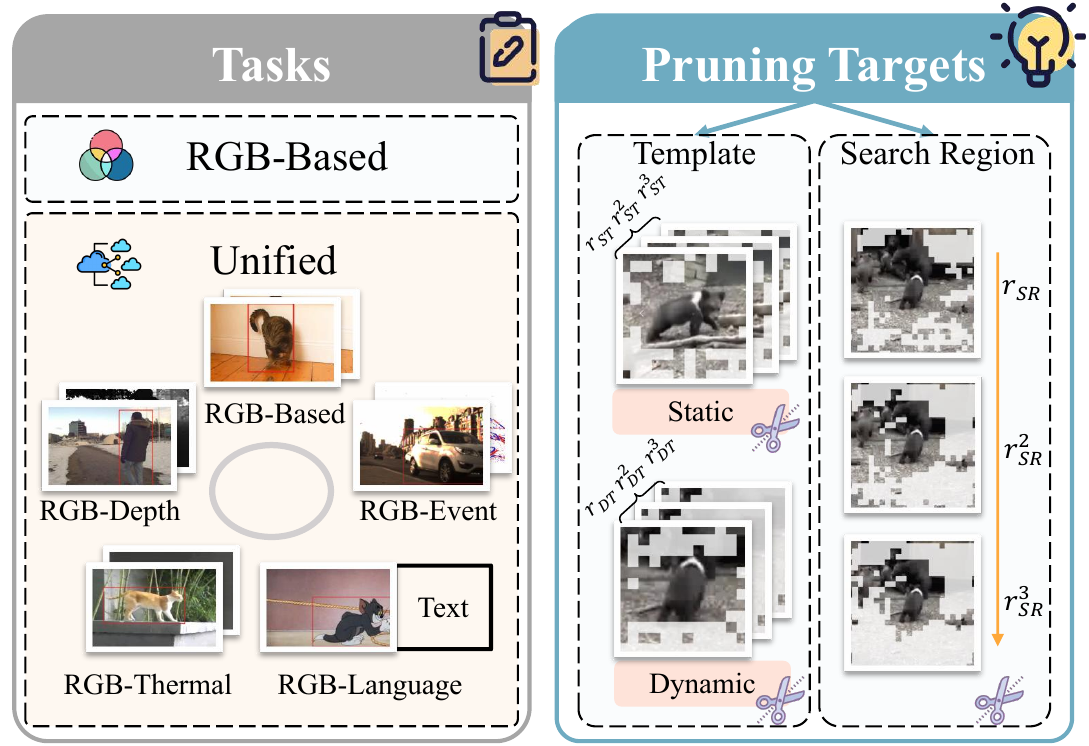}
    \caption{UTPTrack supports RGB-based and unified tracking, prunes redundant tokens in the search region (SR), dynamic template (DT), and static template (ST) to improve efficiency, with $r$ indicating the token retention ratio for each component.}
    \label{fig:first}
    \vspace{-10pt}
\end{figure}

Transformer-based trackers generally categorized as two-stream or one-stream. Two-stream trackers~\cite{chen2021transformer, chen2022efficient, zhu2025two} process the template and search region separately, allowing template reuse and reducing inference cost. In contrast, one-stream trackers~\cite{ye2022joint, cui2022mixformer, chen2022backbone, xie2022correlation} jointly encode both inputs within a unified Transformer, enabling richer template–search interactions and stronger global feature representations. While one-stream trackers have become the dominant paradigm due to their superior performance, they come with significant computational overhead: the quadratic complexity of Transformers, compounded by the large number of video tokens, makes real-time deployment particularly challenging on resource-constrained devices. 

Token pruning has recently improved tracking efficiency by reducing tokens in the search region or the dynamic template~\cite{ye2022joint, lan2023procontext, zhang2025atptrack}. However, current methods underexploit their potential since none performs joint pruning across all three critical components: SR, DT, and ST. Treating these components in isolation overlooks their inherent interdependencies, which are essential for precise localization and boundary-aware tracking. Without a unified redundancy modeling strategy that captures cross-component relationships, existing approaches often make suboptimal decisions: informative tokens may be discarded, and uneven redundancy across components remains underexploited. Consequently, the tracking performance is often compromised through degraded spatial consistency and semantic completeness, a problem that becomes particularly severe in multimodal scenarios where aligning information is critical.

To address this problem, we propose \emph{UTPTrack}, a simple and Unified Token Pruning framework for efficient and accurate tracking (\cref{fig:first}). Unlike previous methods that prune tokens from a single component or rely on heuristic rules, \emph{UTPTrack} jointly prunes all three sources (search region, dynamic template, and static template) within a one-stream architecture, guided by a holistic redundancy modeling strategy. Specifically, we identify redundancy patterns within each component and tailor attention-guided pruning strategies accordingly. For the search region and dynamic template, token relevance is computed based on similarity to the static template center token, retaining only the most important tokens. For the static template, we further enhance pruning robustness via a token type-aware strategy, leveraging spatial priors from the target bounding box to avoid discarding foreground tokens.

Beyond RGB-based tracking, UTPTrack naturally extends to unified tracking, which supports multimodal and language-guided tasks in one model. For additional visual modalities (\eg, depth, thermal, event), pruning is performed in a shared embedding space using the same attention-based mechanism. For RGB–language tasks, we introduce a text-guided pruning strategy, where semantic cues from language tokens jointly guide token selection alongside visual features. This unified and modality-aware design allows UTPTrack to generalize across diverse tracking scenarios while maintaining high efficiency.

We comprehensively evaluate the effectiveness of UTPTrack on OSTrack~\cite{ye2022joint} and SUTrack~\cite{chen2025sutrack} across ten RGB-based and multimodal tracking benchmarks. Extensive results demonstrate its strong generalization and efficiency across diverse tasks. Specifically, UTPTrack reduces token number by 65.4\% on OSTrack\textsubscript{384} and 67.5\% on SUTrack\textsubscript{384}, with corresponding MAC reductions of 31.3\% and 28.4\%, while maintaining or even slightly improving accuracy (99.7\% and 100.5\% of baseline performance).
The contributions of this work are summarized as follows:
\begin{itemize}
\item We introduce \emph{UTPTrack}, the first unified token pruning framework that jointly compresses the SR, DT, and ST within a one-stream Transformer.
\item We propose an \emph{attention-guided, token type-aware strategy} using cross-component similarity and spatial priors to remove redundancy while preserving key information.
\item We extend our framework to \emph{unified tracking} via a unified, modality-aware pruning mechanism, including a novel text-guided strategy that integrates semantic cues from natural language.
\item Extensive experiments across ten benchmarks demonstrate that UTPTrack significantly reduces token count and computation while maintaining state-of-the-art performance in both RGB-based and unified tracking.
\end{itemize}

\section{Related Work}
\label{sec:app_rel}
\paragraph{RGB-based and Unified Object Tracking.}
RGB-based object tracking focuses on a single target with RGB images and underpins many advances. Recently, one-stream Transformer trackers couple feature extraction and relation modeling for stronger representations and more accurate localization~\cite{cui2022mixformer, ye2022joint, chen2022backbone, he2023target}. Building on this flexibility, research has moved to unified frameworks handling multiple modalities (RGB+Depth/Thermal/Event/Language). Parameter-efficient adaptation~\cite{zhu2023visual, hong2024onetracker, hou2024sdstrack} keeps a strong RGB foundation while injecting modality cues via lightweight prompts or adapters. UnTrack~\cite{wu2024single} learns a shared low-rank latent space across modalities, whereas SUTrack~\cite{chen2025sutrack} consolidates five tasks within a single one-stream model via a unified modality representation. For efficiency, some work reduces redundancy through shared backbones, compact experts, distillation~\cite{liu2024emtrack}, or learnable interaction tokens~\cite{hu2025adaptive}. Complementary to these directions, we address the remaining token bottleneck with a token-level economy using hierarchical pruning, improving the compute–performance trade-off without modifying unified architectures.



\begin{figure*}[t!]
    \centering
    \includegraphics[width=\linewidth]{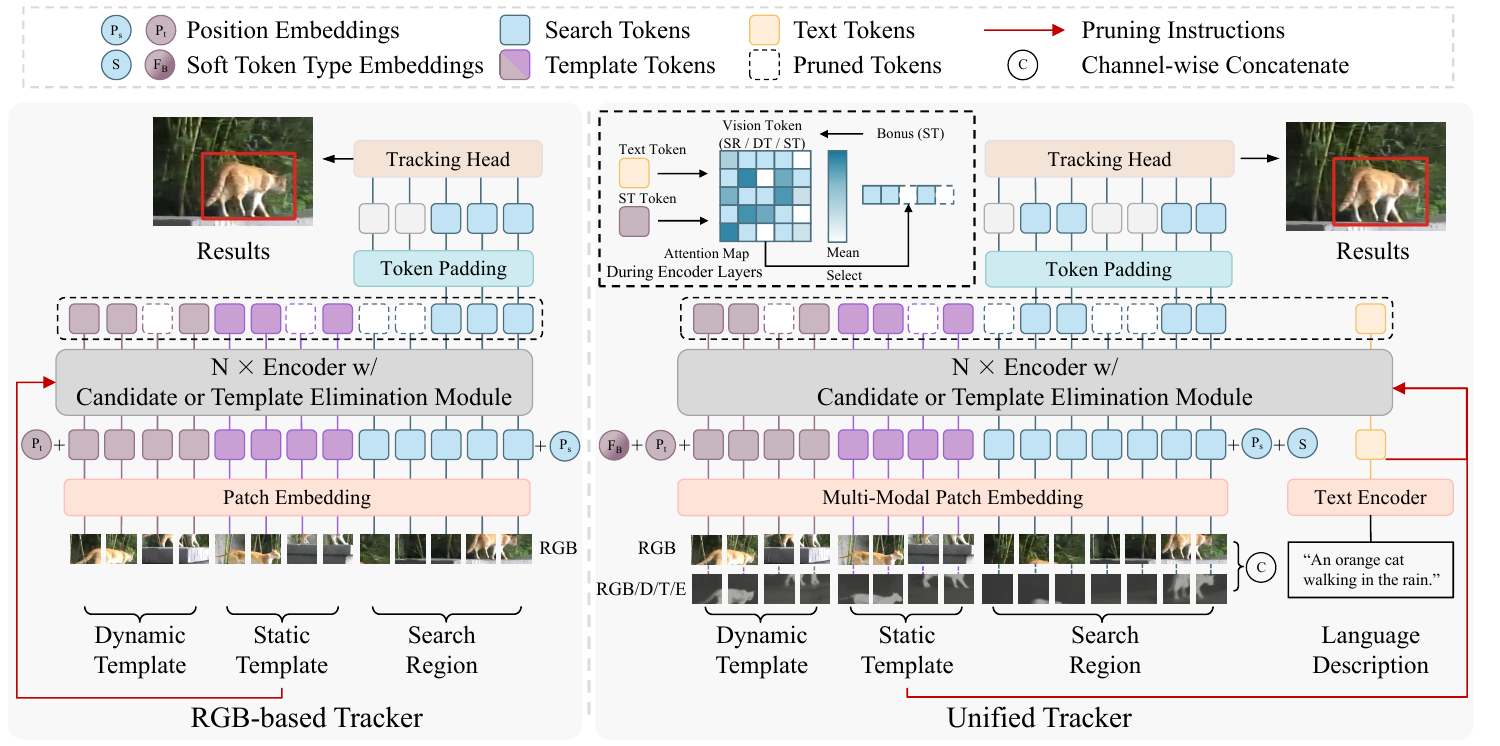} 
    \caption{Architecture of the proposed UTPTrack. UTPTrack supports both RGB-based and unified tracking. It adopts a one-stream transformer that jointly processes tokens from the search region (SR), dynamic template (DT), and static template (ST). A lightweight Candidate or Template Elimination Module (CTEM) is inserted into encoder layers to prune redundant tokens from all three sources. In the figure, D/T/E denote depth, thermal, and event modalities, respectively.}
    \label{fig:utptrack}
    \vspace{-10pt}
\end{figure*}

\vspace{-5pt}
\paragraph{Token Compression in Computer Vision.}
Token compression in ViT comprise two families: pruning and merging. \textbf{Pruning} removes low-importance tokens~\cite{fan2025visipruner}. DynamicViT~\cite{rao2021dynamicvit} predicts token saliency with an MLP and discards the rest. EViT~\cite{liang2022not} and SPViT~\cite{kong2022spvit} estimate importance and either drop or collapse low-saliency tokens before the next layer. Evo-ViT~\cite{xu2022evo} uses evolving CLS attention with slow–fast updates to preserve structure. \textbf{Merging} combines similar tokens. ToMe~\cite{bolya2022token} performs bipartite soft matching to merge a fixed quota of similar tokens. VoMix~\cite{peng2024vote} redistributes redundant token content to key tokens via similarity-weighted mixing. ATM~\cite{lee2025lossless} adopts a layer-wise decreasing similarity threshold and a merging-aware matching strategy. 
Despite extensive study in ViTs, token compression for tracking remains underexplored. OSTrack~\cite{ye2022joint} proposes early elimination of search tokens based on similarity to the central static template. ProContEXT~\cite{lan2023procontext} refines this by incorporating both static and dynamic templates to compute similarity scores to improve robustness. However, these methods prune components in isolation, overlooking redundancy overlap and cross-dependencies among the search region, dynamic template, and static template. In contrast, our method introduces a joint pruning strategy across all three components, aiming to model redundancy holistically and improve pruning effectiveness in more complex tracking scenarios.
\section{Methodology}
\label{sec:methodology}
In this section, we present UTPTrack. We first describe the one-stream pipelines for RGB-based and unified tracking, together with token-pruning preliminaries, then analyze pruning targets (\ie, SR, DT, ST). We further introduce a token type-aware strategy for ST and extend the framework to unified tracking with a modality-aware pruning scheme.

\subsection{Preliminary}
\paragraph{RGB-based Tracking Pipeline.}
\label{sec:preliminary}
Given a ST $\mathbf{Z_s} \in \mathbb{R}^{H_z \times W_z \times 3}$ and the current frame's SR $\mathbf{X} \in \mathbb{R}^{H_x \times W_x \times 3}$, the tracker aims to estimate the target bounding box $\mathbf{B} \in \mathbb{R}^4$. To handle appearance changes, a DT $\mathbf{Z_d} \in \mathbb{R}^{H_z \times W_z \times 3}$ is often maintained. A typical one-stream RGB tracker uses a backbone $\mathcal{F}(\cdot)$ for feature extraction and relation modeling, followed by a tracking head $\varphi(\cdot)$. The inputs are split into patches, embedded, and flattened into token sequences: $\mathbf{E_{x}} \in \mathbb{R}^{N_{x} \times D}$, $\mathbf{E_{sz}} \in \mathbb{R}^{N_{sz}\times D}$, and $\mathbf{E_{dz}} \in \mathbb{R}^{N_{dz} \times D}$. These are concatenated and processed by the backbone:
$\mathbf{B} = \varphi\left( \mathcal{F}(\text{Concat}(\mathbf{E_x}, \mathbf{E_{sz}}, \mathbf{E_{dz}})) \right)$.

\vspace{-10pt}
\paragraph{Unified Tracking Pipeline.}
Unified trackers (\eg, SUTrack) extend RGB trackers by integrating complementary modalities (\eg., depth, thermal, event, language) into a single framework. For depth, thermal, and event, three additional channels (D/T/E) are concatenated with RGB to form a six-channel input. The multimodal features are aligned and projected into unified embeddings, yielding SR $\mathbf{E_{x}} \in \mathbb{R}^{N_{x}\times 2D}$, ST $\mathbf{E_{sz}} \in \mathbb{R}^{N_{sz}\times 2D}$ and DT $\mathbf{E_{dz}} \in \mathbb{R}^{N_{dz}\times 2D}$. For language, a single token is extracted via CLIP-L~\cite{radford2021learning} from the text description and concatenated with visual tokens, enabling cross-modal interaction. The output is computed as:
$ \mathbf{B} = \varphi\left( \mathcal{F}(\text{Concat}(\mathbf{E_x}, \mathbf{E_{sz}}, \mathbf{E_{dz}}, \mathbf{E_{text}})) \right). $
Missing modalities are handled by duplicating RGB channels (for D/T/E) or using a fixed dummy sentence (for text), ensuring consistent input format.

As shown in \cref{fig:utptrack}, both pipelines share a similar structure, with the unified version extending the input for multimodal data while remaining compatible with RGB.

\vspace{-10pt}
\paragraph{Token Pruning in Tracking.}
In ViT~\cite{dosovitskiy2020image}, MHA and MLP process all tokens, incurring $\mathcal{O}(N^2)$ and $\mathcal{O}(N)$ costs, which hinders real-time tracking. Token pruning reduces compute by keeping salient tokens, but tracking requires spatial coherence. Therefore, we restore retained tokens to their original indices and zero-pad pruned slots before the tracking head, preserving spatial layout while reducing computation.


\subsection{UTPTrack}
The architecture of UTPTrack is shown in \cref{fig:utptrack}. Based on the RGB and unified pipelines (\cref{sec:preliminary}), it adopts a one-stream design that jointly processes SR, ST, and DT tokens for tracking. Following OSTrack~\cite{ye2022joint}, we insert a lightweight Candidate or Template Elimination Module (CTEM) into selected layers to prune redundant tokens using attention-derived importance scores. Pruning is guided by the ST in RGB tracking, and additionally by language in unified tracking. To maintain spatial alignment, pruned SR tokens are zero-padded to their original positions before the tracking head. Details of CTEM are provided in \cref{sec:pruning}.

\subsubsection{Token Pruning for RGB-based Tracking}
\label{sec:pruning}

In the one-stream architecture, SR, ST, and DT tokens are jointly processed via attention layers. While this enables rich cross-token interactions, it also introduces redundancy, especially from background regions with limited discriminative value. To mitigate this, we design dedicated elimination modules to prune unimportant tokens from each source.

Pruning is guided by attention-based similarity, reusing attention weights from the transformer encoder to measure token relevance without extra computation. Given the concatenated token sequence $[\mathbf{E_x}; \mathbf{E_{sz}}; \mathbf{E_{dz}}]$ from $\mathbf{X}$, $\mathbf{Z_s}$, and $\mathbf{Z_d}$, the attention is computed as:
{\small
\setlength{\arraycolsep}{2pt}
\begin{equation}
\begin{aligned}
    & \text{Attention}(Q, K, V) = \text{Softmax}\left( \frac{QK^T}{\sqrt{d_k}} \right) V \\ 
    = & \text{Softmax} \left( \frac{[Q_x; Q_{sz}; Q_{dz}][K_x; K_{sz}; K_{dz}]^T}{\sqrt{d_k}} \right)
    \begin{bmatrix}
        V_x \\
        V_{sz} \\
        V_{dz}
    \end{bmatrix}
    \label{eq:attn}
\end{aligned}
\end{equation}
}
where $Q$, $K$, and $V$ denote query, key, and value matrices, with subscripts $x$, $sz$, $dz$ indicating the SR, ST, and DT, respectively. The attention weights $A$ in \cref{eq:attn} expand to:
{\small
\setlength{\arraycolsep}{2pt}
\begin{equation}
A = \text{Softmax} \left( \frac{1}{\sqrt{d_k}} 
\begin{bmatrix}
Q_x K_x^T & Q_x K_{sz}^T & Q_x K_{dz}^T \\
Q_{sz} K_x^T & Q_{sz} K_{sz}^T & Q_{sz} K_{dz}^T \\
Q_{dz} K_x^T & Q_{dz} K_{sz}^T & Q_{dz} K_{dz}^T
\end{bmatrix}
\right)
\label{eq:attn_map_rgb}
\end{equation}
}
This dense interaction benefits feature fusion but risks noise propagation. For instance, terms like $Q_x K_x^T$, $Q_x K_{sz}^T$, and $Q_x K_{dz}^T$ may allow background tokens to attend to each other or dominate target-related responses. Similarly, template tokens attending to noisy search tokens (\eg., via $Q_{sz} K_x^T$, $Q_{dz} K_x^T$) can degrade representation quality.

\vspace{-5pt}
\paragraph{Candidate (Search Region) Elimination.}
The SR tokens often contain background clutter, introducing redundancy and harming localization accuracy. This is evident in attention terms like $Q_x K_x^T$, where background tokens attend to each other, and $Q_x K_{sz}^T$ or $Q_x K_{dz}^T$, where SR tokens attend to non-discriminative template regions. To address this, we measure the importance of each SR token based on its attention similarity to the ST. According to \cref{eq:attn_map_rgb}, the relevance score is computed by $\omega_x =\text{softmax}(Q_{sz'} K_x^T / \sqrt{d_k}) \in \mathbb{R}^{1\times N_x}$, where $Q_{sz'}$ is the query of the ST’s center token and $N_x$ is the number of SR tokens. We retain the top-$k$ tokens and prune the rest, reducing background interference while preserving target-relevant information.


\vspace{-5pt}
\paragraph{Dynamic Template Elimination.}
The DT may contain noisy tokens due to drift, occlusion, or appearance changes, which can negatively affect the SR and ST via attention terms like $Q_{dz}K^T_x$ and $Q_{dz}K^T_{sz}$. Similar to the SR, we compute the similarity between each DT token and the ST's center token as $\omega_{dz} =\text{softmax}(Q_{sz'} K_{dz}^T / \sqrt{d_k}) \in \mathbb{R}^{1\times N_{dz}}$,  and pruning those with low relevance.

\vspace{-5pt}
\paragraph{Static Template Elimination.}
Since the ST is obtained by enlarging the target box, it may include background tokens that introduce irrelevant attention interactions (e.g., $Q_{sz}K^T_x$ and $Q_{sz}K^T_{dz}$). To refine its representation, we measure each token’s similarity to the center token as
$\omega_{sz} = \text{softmax}(Q_{sz'}K_{sz}^T / \sqrt{d_k}) \in \mathbb{R}^{1\times N_{sz}}$,
and prune tokens with low relevance, while always preserving the center token.

\vspace{-5pt}
\paragraph{Token Type Aware Pruning.}
\label{sec:tta}
Motivated by the token-type embedding mechanisms~\cite{lin2024tracking, chen2025sutrack}, we propose a token type-aware pruning strategy during ST elimination to suppress the probability of mistakenly discarding foreground tokens. Given a ST $\mathbf{Z_s} \in \mathbb{R}^{H_z \times W_z \times 3}$ and its bounding box $\mathbf{B}$, we construct a binary mask $\mathbf{M} \in \mathbb{R}^{H_z \times W_z}$, where pixels inside $\mathbf{B}$ are set to 1 and those outside to 0:
{\small
\begin{equation}
M(i,j) = 1 \text{ if } (i,j)\text{ is inside } B,\; 0 \text{ otherwise.}
\end{equation}
}
We divide the mask into non-overlapping $P^2$ patches $\mathbf{M}^{(k)}_{\text{patch}}$ and assign each patch a foreground score by averaging its mask values. This score serves as a bonus to encourage retaining foreground tokens during pruning and is incorporated into the attention-based elimination module (see \cref{fig:utptrack}). We explore three bonus strategies to control how this foreground prior influences pruning.

\begin{itemize}
    \item Full bonus assigns 1 if all pixels lie inside the bounding box. The value for patch $\mathbf{M}^{(k)}_{\text{patch}}$ is:
\end{itemize}
{\small 
\begin{equation}
    b^{(k)}_{\text{full}} = 1 \text{if } \mathbf{M}(i,j) = 1 \text{ for all } (i,j) \in \mathbf{M}^{(k)}_{\text{patch}}, \; 0 \text{ otherwise}.
\end{equation}
}
\begin{itemize}
    \item Soft bonus uses the mean mask value of each patch:
\end{itemize}
{\small 
\begin{equation}
    b^{(k)}_{\text{soft}} = m^{(k)}_{\text{avg}}
    = \frac{1}{P^2} \sum \mathbf{M}(i,j),
    \quad (i,j) \in \mathbf{M}^{(k)}_{\text{patch}} .
\end{equation}
}
\begin{itemize}
    \item All bonus assigns 1 if any pixel lies inside the box:
\end{itemize}
{\small
\begin{equation}
  b^{(k)}_{\text{all}} = 1 \text{ if } \exists (i,j)\in \mathbf{M}^{(k)}_{\text{patch}}
  \text{ : } \mathbf{M}(i,j)=1,\; 0 \text{ otherwise}.
\end{equation}
}
These bonuses are added directly to attention scores during ranking to guide pruning toward foreground-relevant tokens. As shown in \cref{fig:utptrack}, this mechanism is integrated into the elimination module. The soft version is used by default.

\subsubsection{Token Pruning for Unified Tracking}
The proposed pruning strategy, originally designed for RGB tracking, naturally extends to multimodal settings. For visual modalities (RGBD, RGBT, RGBE), auxiliary channels (depth, thermal, event) are concatenated with RGB to form a six-channel input. After projection into a unified embedding space, token dimensions increase, but their spatial layout remains unchanged. Thus, pruning based on attention to the ST's center token remains effective without modification. For language modality, we additionally account for semantic cues: the text description is encoded into a single token using CLIP-L~\cite{radford2021learning} and concatenated with visual tokens before the transformer, extending the attention in \cref{eq:attn_map_rgb} to include interactions with the language token. The updated attention weights $A$ are:
{\small
\setlength{\arraycolsep}{2pt}
\begin{equation}
A\!\!=\!\! \text{Softmax} \!\!
\left( 
\!\!\frac{1}{\sqrt{d_k}}
\!\!
\begin{bmatrix}
    Q_x K_x^T & Q_x K_{sz}^T & Q_x K_{dz}^T & Q_x K_{t}^T \\
    Q_{sz} K_x^T & Q_{sz} K_{sz}^T & Q_{sz} K_{dz}^T & Q_{sz} K_{t}^T \\
    Q_{dz} K_x^T & Q_{dz} K_{sz}^T & Q_{dz} K_{dz}^T & Q_{dz} K_{t}^T \\
    Q_{t} K_x^T & Q_{t} K_{sz}^T & Q_{t} K_{dz}^T & Q_{t} K_{t}^T \\
\end{bmatrix}
\! \right)\!
\label{eq:attn_map_unified}
\end{equation}
}
\paragraph{Text-Guided Pruning.}
Based on the updated attention in \cref{eq:attn_map_unified}, the language token interacts with all visual tokens through bidirectional attention, enabling semantic cues to enhance spatial representations. To leverage this for pruning, we jointly use the ST's center token and the language token to guide token importance estimation. Specifically, the importance of each token is computed as:
{\small
\begin{equation}
\omega_x = \phi \left(
\text{softmax}\left( \frac{Q_{sz'} K_x^T}{\sqrt{d_k}} \right)
+
\text{softmax}\left( \frac{Q_{t} K_x^T}{\sqrt{d_k}} \right)
\right)
\label{eq:text_guided_pruning}
\end{equation}
}
where $\phi(\cdot)$ denotes summation across attention maps, and $Q_t$ is the query of the text token. This enables pruning to benefit from both spatial and semantic guidance. The same principle can applie to the DT and ST. Related ablations are provided in \cref{sec:ablation}, where we study which token group benefit most from text-guided pruning.

\subsection{Training and Inference}
Following OSTrack~\cite{ye2022joint}, our RGB-based trackers use a weighted focal loss for classification, L1 and generalized IoU losses for bounding box regression. The total loss is:
{\small
\begin{equation}
    \mathcal{L}_{\text{RGB}} = \lambda_{\text{cls}}\mathcal{L}_{\text{cls}} + \lambda_{\text{giou}} \mathcal{L}_{\text{giou}} + \lambda_{L_1} \mathcal{L}_{L_1},
\end{equation}
}
where $\mathcal{L}_{\text{cls}}$, $\mathcal{L}_{\text{giou}}$, and $\mathcal{L}_{L_1}$ denote the weighted focal, generalized IoU, and L\textsubscript{1} losses. We set $\lambda_{\text{cls}} = 1$, $\lambda_{\text{giou}} = 2$, and $\lambda_{L_1} = 5$. For unified tracking, following SUTrack~\cite{chen2025sutrack}, we incorporate a cross-entropy loss for task recognition:
{\small
\begin{equation}
    \mathcal{L}_{\text{Unified}} = \mathcal{L}_{\text{RGB}} + \lambda_{\text{task}} \mathcal{L}_{\text{task}},
\end{equation}
}
where $\mathcal{L}_{\text{task}}$ is the task-recognition cross-entropy loss, with $\lambda_{\text{task}} = 1$. During inference, RGB-based and unified trackers adopt a dual-template strategy using one ST and one DT. The DT is updated based on a fixed time interval and confidence threshold, following the strategy of STARK~\cite{yan2021learning}.


\section{Experiment}
\subsection{Implementation Details}


\paragraph{Model.}
We present four UTPTrack variants across architectures and resolutions for RGB and unified tracking. Model names encode the search region resolution (number) and base tracker (suffix). OSTrack~\cite{ye2022joint} is a modified version without the candidate elimination module and with a single DT. \Cref{tab:details} reports parameters, MACs, and inference speed. The negative MAC subscript indicates computation reduction and the FPS subscript denotes the speed gain from pruning, while the parameter subscript marks the CLIP-L~\cite{radford2021learning} text encoder used in language-guided tracking.

\begin{table}[t!]
    \vspace{-5pt}
    \caption{Details of UTPTrack model variants.}
    \label{tab:details}
    \centering
    \resizebox{\columnwidth}{!}{
    \begin{tabular}{@{}c|ccccc@{}}
        \toprule
        Method & 
        \makecell{Base\\Model} & 
        \makecell{Params\\(M)} & 
        \makecell{MACs\\(G)} & 
        \makecell{FPS\\(GPU)} &
        \makecell{FPS\\(CPU)} \\
        \midrule
        UTPTrack-O\textsubscript{256}  & OSTrack-256 & 92 & 24 (-11) & 95 (+1) & 17 (+8) \\
        UTPTrack-O\textsubscript{384}  & OSTrack-384 & 92 & 53 (-25) & 47 (+7) & 6 (+3) \\
        UTPTrack-S\textsubscript{224} & SUTrack-B224 & 70 (+85) & 16 (-7) & 43 (+2) & 9 (+1) \\
        UTPTrack-S\textsubscript{384} & SUTrack-B384 & 70 (+85) & 48 (-19) & 31 (+4) & 5 (+2) \\
        \bottomrule
    \end{tabular}
    }
    \vspace{-10pt}
\end{table}

\vspace{-10pt}
\paragraph{Training.}
Following standard practice in RGB-based tracking~\cite{ye2022joint}, UTPTrack-O are trained on TrackingNet~\cite{muller2018trackingnet}, LaSOT~\cite{fan2019lasot}, GOT-10k~\cite{huang2019got}, and COCO~\cite{lin2014microsoft} for 300 epochs with 60k image pairs per epoch. For unified tracking, UTPTrack-S follows the SUTrack and uses a broader multimodal datasets, including TrackingNet~\cite{muller2018trackingnet}, LaSOT~\cite{fan2019lasot}, GOT-10k~\cite{huang2019got}, COCO~\cite{lin2014microsoft}, TNL2K~\cite{wang2021towards}, VASTTrack~\cite{peng2024vasttrack}, DepthTrack~\cite{yan2021depthtrack}, LasHeR~\cite{li2021lasher}, and VisEvent~\cite{wang2023visevent}, trained for 180 epochs with 100k image pairs per epoch. For all variants, template and search region crops are generated by expanding the target bounding box by 2$\times$ and 4$\times$, respectively. Training is conducted on 4 NVIDIA A100 GPUs. 

\vspace{-10pt}
\paragraph{Inference.}
During inference, the DT is updated every 25 frames using a confidence threshold of 0.7. A Hanning window penalty is applied to incorporate position priors, following common practice~\cite{ye2022joint, chen2025sutrack}. Inference speed is evaluated on a single NVIDIA 1080 Ti GPU and an Intel Xeon Gold 6226R @ 2.90GHz CPU.

\subsection{State-of-the-Art Comparisons}
We verify the effectiveness of UTPTrack under two tracking paradigms: (1) RGB-based tracking and (2) Unified tracking, which includes RGB-based, RGB-Depth, RGB-Thermal, RGB-Event, and RGB-Language tasks. For each paradigm, we use a representative base model (OSTrack or SUTrack) and compare UTPTrack against state-of-the-art token compression methods, including CE~\cite{ye2022joint}, ToMe~\cite{bolya2022token}, EViT~\cite{liang2022not}, and DynamicViT~\cite{rao2021dynamicvit}, under varying input resolutions. We report results under two complementary protocols: (1) Overall Performance, using released default settings, and (2) Controlled-Budget Performance, evaluated at  three matched token compression ratio.    

\begin{figure}[t!]
    \centering
    \includegraphics[width=\linewidth]{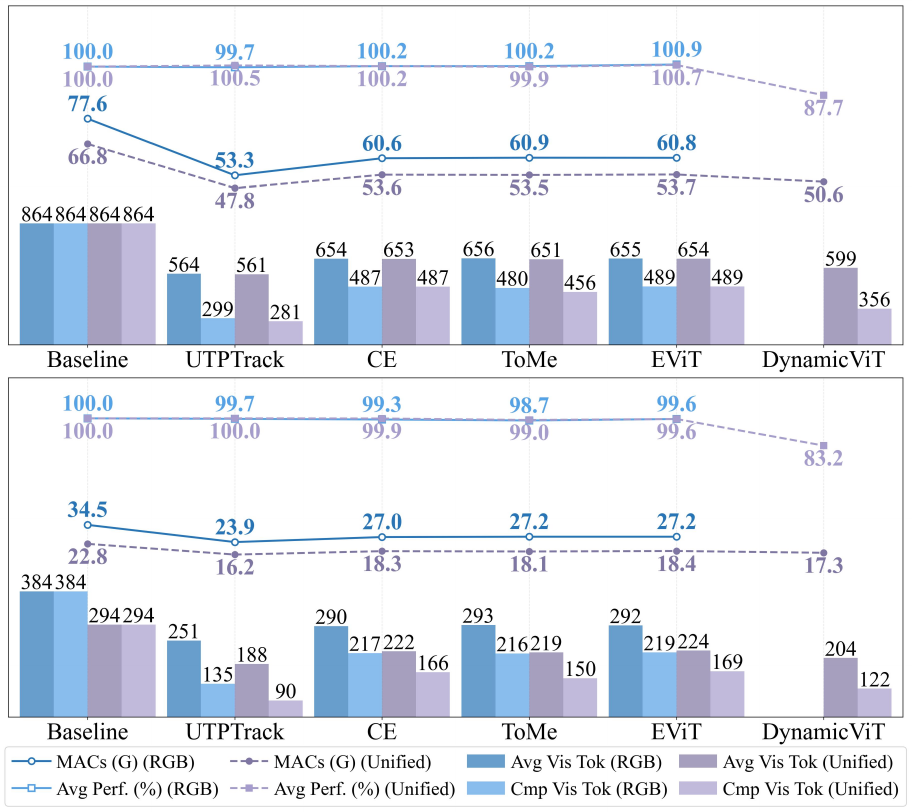}
    \caption{Performance comparison of UTPTrack and other pruning methods under each method's default compression settings at two resolutions. Top (High Resolution): 384 (RGB and Unified). Bottom (Low Resolution): 256 (RGB), and 224 (Unified).}
    \label{fig:overall_result}
    \vspace{-14pt}
\end{figure}


\vspace{-15pt}
\paragraph{Overall Performance.} \label{sec:overall_exp}
As shown in \cref{fig:overall_result}, UTPTrack achieves the best performance-efficiency trade-off across both RGB-based and Unified tracking settings. 
At the low-resolution setting, it attains the highest compression ratio while delivering the best performance in both paradigms. Specifically (RGB-based / Unified), MACs drop by 30.7\% (34.5G $\to$ 23.9G) / 28.9\% (22.8G $\to$ 16.2G) and vision tokens prune by 64.8\% (384 $\to$ 135) / 69.4\% (294 $\to$ 90), while maintaining 99.7\% / 100.0\% of baseline performance. Under comparable performance, it further achieved a 1.5$\times$ / 1.6$\times$ compression. 
At the high-resolution setting, UTPTrack sustains its advantage and scales favorably: (RGB-based / Unified) MACs are reduced by 31.3\% / 28.4\%, tokens by 65.4\% / 67.5\%, with performance preserved at 99.7\% / 100.5\% of the baseline. Notably, as the resolution increases, UTPTrack-S even surpasses the baseline, reaching 100.5\% performance. This small but consistent gain suggests that our token selection acts as a mild regularizer by removing redundant or noisy visual tokens and sharpening attention to salient regions, thereby improving accuracy while reducing computation. Comprehensive per-benchmark and per-metric results are provided in \cref{app:overall_performance}, with detailed practical efficiency analyses in \cref{app:efficiency}.
Unlike methods such as ToMe and DynamicViT that rely on external heuristics or auxiliary prediction modules, UTPTrack directly utilizes the model’s inherent attention maps to guide token pruning. This internal mechanism preserves alignment with task-specific semantics while introducing minimal architectural overhead. Movever, it is lightweight and architecture-agnostic, making it compatible with any transformer-based tracker without the need for retraining or structural modifications.

\begin{table}[t!]
    \caption{Performance comparisions under different vision token configurations across \textbf{RGB-based tracking}. All methods are applied on the same base model \textbf{OSTrack\textsubscript{256}}. The average performance listed is calculated across all four benchmarks.}
    \label{tab:budget_rgb}
    \centering
    \resizebox{\linewidth}{!}{
    \begin{tabular}{l|cccc|c}
        \toprule
        \textbf{Method} & 
        \rotatebox{75}{\textbf{LaSOT}} & 
        \rotatebox{75}{\textbf{LaSOT\textsubscript{ext}}} & 
        \rotatebox{75}{\textbf{TrkNet}} & 
        \rotatebox{75}{\textbf{GOT-10k}} & 
        \textbf{Avg Perf. (\%)} \\
        \midrule
        \rowcolor{gray!20}
        \multicolumn{6}{c}{\textit{Upper Bound, 384 Tokens (100\%)}} \\
        \midrule
        OSTrack$_{256}$ (Baseline) & 67.9 & 47.6 & 84.5 & 74.9 & 100\% \\
        \midrule
        \rowcolor{gray!20}
        \multicolumn{6}{c}{\textit{Retain 87.2\% Tokens in Average ($\approx$ 335 tokens, \textcolor{ForestGreen}{$\downarrow$ 18.8\%)}}} \\
        \midrule
        OSTrack-CE\textsubscript{256}           & \underline{67.1} & \underline{46.9} & \textbf{84.1} & 74.9 & \textbf{99.3\%} ($\downarrow0.7\%$) \\
        OSTrack-ToMe\textsubscript{256}         & \textbf{68.0} & 46.4 & 83.8 & 74.3 & 99.0\% ($\downarrow1.0\%$) \\
        OSTrack-EViT\textsubscript{256}         & 66.8 & 46.7 & \textbf{84.1} & \underline{75.0} & 99.0\% ($\downarrow1.0\%$) \\
        UTPTrack-O\textsubscript{256} (ours)    & \underline{67.1} & \textbf{48.3} & \underline{84.0} & \textbf{75.6} & \textbf{100.2\%} ($\uparrow0.2\%$) \\
        \midrule
        \rowcolor{gray!20}
        \multicolumn{6}{c}{\textit{Retain 75.5\% Tokens in Average ($\approx$ 290 tokens, \textcolor{ForestGreen}{$\downarrow$ 24.5\%)}}} \\
        \midrule
        OSTrack-CE\textsubscript{256}           & 67.3 & \textbf{47.3} & \textbf{84.2} & 74.2 & 99.3\% ($\downarrow0.7\%$) \\
        OSTrack-ToMe\textsubscript{256}         & \textbf{68.2} & 46.5 & \textbf{84.2} & \textbf{74.9} & \underline{99.4\%} ($\downarrow0.6\%$) \\
        OSTrack-EViT\textsubscript{256}         & \underline{67.8} & \underline{47.1} & \underline{83.8} & 72.9 & 98.8\% ($\downarrow1.2\%$) \\
        UTPTrack-O\textsubscript{256} (ours)    & 67.6 & \textbf{47.3} & \underline{83.8} & \underline{74.8} & \textbf{99.5\%} ($\downarrow0.5\%$) \\
        \midrule
        \rowcolor{gray!20}
        \multicolumn{6}{c}{\textit{Retain 65.6\% Tokens in Average ($\approx$ 252 tokens, \textcolor{ForestGreen}{$\downarrow$ 34.4\%)}}} \\
        \midrule
        OSTrack-CE\textsubscript{256}           & 67.2 & \textbf{46.5} & \underline{83.3} & \underline{72.8} & \underline{98.1\%} ($\downarrow1.9\%$) \\
        OSTrack-ToMe\textsubscript{256}         & 66.3 & 46.0 & 82.5 & 71.0 & 96.7\% ($\downarrow3.3\%$) \\
        OSTrack-EViT\textsubscript{256}         & \underline{68.1} & 46.1 & 83.0 & \underline{72.8} & \underline{98.1\%} ($\downarrow1.9\%$) \\
        UTPTrack-O\textsubscript{256} (ours)    & \textbf{68.2} & \underline{46.2} & \textbf{84.0} & \textbf{74.9} & \textbf{99.2\%} ($\downarrow0.8\%$) \\
        \bottomrule
    \end{tabular}
    }
    \vspace{-10pt}
\end{table}

\begin{table*}[t!]
    \caption{Performance comparisons under different vision token compression configurations across \textbf{unified tracking}. All methods are applied on the same base model \textbf{SUTrack\textsubscript{224}}. The best result are \textbf{bolded} and the second best results are \underline{underlined} in all following tables. The average performance listed is calculated across all 10 benchmarks. TrackingNet is abbreviated as TrkNet for brevity.}
    \label{tab:budget_unified}
    \centering
    \resizebox{\linewidth}{!}{
    \scriptsize
    \begin{tabular}{l|cccc|c|cc|c|cc|c}
        \toprule
        & \multicolumn{4}{c|}{\textbf{RGB}} & \textbf{RGB-D} & \multicolumn{2}{c|}{\textbf{RGB-T}} & \textbf{RGB-E} & \multicolumn{2}{c|}{\textbf{RGB-Lang}} &  \\
        \textbf{Method} & 
        \rotatebox{75}{\textbf{LaSOT}} & 
        \rotatebox{75}{\textbf{LaSOT\textsubscript{ext}}} & 
        \rotatebox{75}{\textbf{TrkNet}} & 
        \rotatebox{75}{\textbf{GOT-10k}} & 
        \rotatebox{75}{\textbf{\makecell{VOT-\\RGBD22}}} &
        \rotatebox{75}{\textbf{LasHeR}} &
        \rotatebox{75}{\textbf{RGBT234}} &
        \rotatebox{75}{\textbf{VisEvent}} &
        \rotatebox{75}{\textbf{TNL2K}} &
        \rotatebox{75}{\textbf{OTB99}} & 
        \textbf{Avg Perf. (\%)}   \\ 
        \midrule
        \addlinespace[0.4ex] \rowcolor{gray!20}
        \multicolumn{12}{c}{\textit{Upper Bound, 294 Tokens (100\%)}} \\ \addlinespace[-0.15ex]
        \midrule
        SUTrack\textsubscript{224} (Baseline) & 73.7 & 53.2 & 85.9 & 77.5 & 75.5 & 59.9 & 70.0 & 63.3 & 67.8 & 68.4  & 100\% \\
        \midrule
        \addlinespace[0.4ex] \rowcolor{gray!20}
        \multicolumn{12}{c}{\textit{Retain 71.4\% Tokens in Average ($\approx$ 218 tokens, \textcolor{ForestGreen}{$\downarrow$ 25.6\%})}} \\ \addlinespace[-0.15ex]
        \midrule
        SUTrack-CE\textsubscript{224}       & \textbf{72.4} & \underline{52.8} & \underline{85.1} & \underline{77.4} & 75.8 & 59.9 & \underline{70.0} & 61.8 & 66.0 & 70.5 & \underline{99.5\%} ($\downarrow0.5\%$) \\
        SUTrack-ToMe\textsubscript{224}     & \underline{72.1} & 52.6 & \textbf{85.4} & 76.9 & \underline{76.0} & \underline{60.0} & 69.0 & \textbf{62.3} & \underline{66.2} & 70.4 & 99.4\% ($\downarrow0.6\%$) \\
        SUTrack-EViT\textsubscript{224}     & 71.6 & 52.3 & 84.9 & 76.9 & 75.7 & 59.0 & \underline{70.0} & \underline{62.0} & 65.7 & \underline{70.7} & 99.0\% ($\downarrow1.0\%$) \\
        SUTrack-DyViT\textsubscript{224}    & 68.8 & 49.7 & 80.2 & 74.6 & 74.4 & 57.6 & 68.7 & 61.3 & 64.0 & \textbf{71.5} & 96.5\% ($\downarrow3.5\%$) \\
        UTPTrack-S\textsubscript{224} (ours)       & \underline{72.1} & \textbf{52.9} & \underline{85.1} & \textbf{77.7} & \textbf{76.5} & \textbf{60.1} & \textbf{70.2} & \underline{62.0} & \textbf{66.4} & \underline{70.7} & \textbf{99.8\%} ($\downarrow0.2\%$) \\
        \midrule
        \addlinespace[0.4ex] \rowcolor{gray!20}
        \multicolumn{12}{c}{\textit{Retain 52.0\% Tokens in Average ($\approx$ 153 tokens, \textcolor{ForestGreen}{$\downarrow$ 48.0\%})}} \\ \addlinespace[-0.15ex]
        \midrule
        SUTrack-CE\textsubscript{224}       & \underline{72.0} &  \underline{52.8} & \underline{85.3} & \underline{77.1} & \textbf{76.1} & 58.4 & \underline{70.4} & 61.5 & \textbf{66.1} & \textbf{70.5} & \underline{99.2\%} ($\downarrow0.8\%$) \\
        SUTrack-ToMe\textsubscript{224}     & 71.4 & 52.0 & 85.2 & \textbf{77.5} & \textbf{76.1} & \underline{59.2} & 69.4 & \underline{61.7} & \underline{65.9} & \textbf{70.5} & 99.0\% ($\downarrow0.5\%$) \\
        SUTrack-EViT\textsubscript{224}     & 71.2 & 51.9 & 84.7 & 76.9 & 75.2 & 57.8 & 69.9 & \textbf{62.3} & 65.1 & \textbf{70.5} & 98.5\% ($\downarrow1.5\%$) \\
        SUTrack-DyViT\textsubscript{224}    & 53.2 & 36.5 & 58.2 & 52.2 & 63.0 & 50.6 & 63.3 & 49.6 & 54.3 & 65.3 & 78.8\% ($\downarrow21.2\%$) \\
        UTPTrack-S\textsubscript{224} (ours)       & \textbf{72.5} & \textbf{53.1} & \textbf{85.5} & \underline{77.1} & \underline{75.7} & \textbf{59.6} & \textbf{70.7} & \underline{61.7} & \underline{65.9} & \underline{70.1} & \textbf{99.5\%} ($\downarrow0.5\%$) \\
        \midrule
        \addlinespace[0.4ex] \rowcolor{gray!20}
        \multicolumn{12}{c}{\textit{Retain 35.4\% Tokens in Average ($\approx$ 104 tokens, \textcolor{ForestGreen}{$\downarrow$ 64.6\%})}} \\ \addlinespace[-0.15ex]
        \midrule
        SUTrack-CE\textsubscript{224}       & \underline{70.5} & \underline{51.8} & \underline{84.8} & \underline{75.9} & \underline{75.5} & \underline{57.8} & \underline{69.3} & \underline{61.4} & \underline{65.4} & \textbf{71.7} & \underline{98.3\%} ($\downarrow1.7\%$) \\
        SUTrack-ToMe\textsubscript{224}     & 67.9 & 48.6 & 80.0 & 71.6 & 70.9 & 53.8 & 63.6 & 58.0 & 61.7 & 67.3 & 92.5\% ($\downarrow7.5\%$) \\
        SUTrack-EViT\textsubscript{224}     & 69.6 & 50.6 & 84.3 & 74.0 & 75.0 & 56.4 & 67.8 & 60.8 & 64.4 & \underline{70.4} & 96.7\% ($\downarrow3.3\%$) \\
        SUTrack-DyViT\textsubscript{224}    & 4.4  & 2.9  & 8.8  & 7.4  & 13.7 & 10.3 & 17.9 & 7.3  & 15.9 & 13.6 & 14.7\% ($\downarrow85.3\%$) \\
        UTPTrack-S\textsubscript{224} (ours)      & \textbf{72.3} & \textbf{52.7} & \textbf{85.2} & \textbf{77.5} & \textbf{76.1} & \textbf{58.6} & \textbf{70.4} & \textbf{61.5} & \textbf{66.0} & 70.3 & \textbf{99.3\%} ($\downarrow0.7\%$) \\
        \bottomrule
    \end{tabular}
    }
    \vspace{-5pt}
\end{table*}

\vspace{-10pt}
\paragraph{Controlled-Budget Performance.}
\label{sec:budget_exp}
To ensure a fair comparison across methods, we conduct controlled-budget experiment under three different compression ratio at low-resolution setting. 
For RGB-based tracking, as shown in \cref{tab:budget_rgb}, we compare UTPTrack-O against other methods across four widely used RGB-based benchmarks. UTPTrack-O consistently outperforms all counterparts at all pruning ratios. Notably, at a moderate budget, UTPTrack-O\textsubscript{256} surpasses the unpruned baseline even after pruning 18.8\% of vision tokens, indicating that a substantial portion of tokens is redundant and that selective pruning can improve generalization.
For Unified tracking, as shown in \cref{tab:budget_rgb},  we compare UTPTrack-S against other methods across elevent widely used RGB-based and multimodal benchmarks. UTPTrack-S\textsubscript{224} likewise maintains a clear lead at every pruning ratios, confirming that the proposed pruning policy transfers effectively beyond RGB-only settings.
Compared with the most similar attention-based token pruning approach, CE~\citep{ye2022joint}, UTPTrack achieves higher performance on nearly all benchmarks under the all three pruning ratios for both RGB-based and Unified Tracking. Notably, as the compression ratio increases, the performance gap widens in favor of UTPTrack. We attribute this to our cross-component joint pruning strategy, which effectively identifies and reduces redundancy across all components, better preserving critical areas. 
The complete per-metrics of each benchmark are provided in \cref{app:budget}.

\subsection{Ablation and Analysis}
\label{sec:ablation}
In this section, we ablate UTPTrack on both RGB-based and unified tracking, using OSTrack\textsubscript{256} and SUTrack\textsubscript{224} as respective baselines. We incrementally analyze the contribution of key components, including Search Region Candidate Elimination (CE), Dynamic Template Elimination (DTE), Static Template Elimination (STE), Token Type-Aware Pruning (TTA), and Text-Guided Pruning (TG). The results demonstrate that our framework effectively reduces token redundancy while preserving tracking accuracy, achieving an efficient performance trade-off.

\vspace{-8pt}
\paragraph{Component Analysis in RGB-based Tracking.}
In \cref{tab:rgb_ablation}, the baseline processes 384 tokens per layer at a cost of 34.5G MACs. With the stepwise addition of CE, DTE, and STE, tokens are reduced to 135 (65\% pruned), and MACs to 23.8G (31\% lower), with 98.9\% performance maintained. Driven by a non-zero pruning budget and strong ST guidance, STE incurs a 0.7\% drop, and TTA, which incorporates bounding-box priors to guide token pruning  within ST, mitigates pruning errors and recovers 0.8\%, reaching 99.7\%. Notably, while DTE and STE (with TTA) further shrink tokens (321→271→135), performance is not monotonically decreasing: DTE yields $+$0.3\% over CE and STE (with TTA) yields $+$0.1\% over DTE, suggesting that pruning redundancy can improve generalization. Overall, the full framework delivers substantial redundancy and computation cuts with minimal loss.

\begin{table}[t!]
    \caption{Ablation Study on \textbf{RGB-based Trackers}. $\Delta$ denotes the averaged performance change from the to row above.}
    \label{tab:rgb_ablation}
    \centering
    \resizebox{\columnwidth}{!}{
    \setlength{\tabcolsep}{4.0pt} 
    \begin{tabular}{@{}c|c|cccc|ccccc@{}}
        \toprule
        \# & Method & 
        \rotatebox{90}{LaSOT} & 
        \rotatebox{90}{LaSOT\textsubscript{ext}} & 
        \rotatebox{90}{TrkNet} & 
        \rotatebox{90}{GOT-10k} & 
        \rotatebox{90}{\makecell{Avg\\Vis Tok}} & 
        \rotatebox{90}{\makecell{Cmp\\Vis Tok}} & 
        \rotatebox{90}{\makecell{MACs\\(G)}} & 
        \rotatebox{90}{\makecell{Avg\\Perf. (\%)}} & 
        $\Delta$ \\
        \midrule
        1 & Baseline    & 67.9 & 47.6 & 84.5 & 74.9 & 384 & 384 & 34.5 & 100.0 & -\\
        2 & + CE        & 67.4 & 46.6 & 83.7 & 75.5 & 291 & 217 & 27.0 & 99.3  & $\downarrow$ 0.7 \\
        3 & + DTE       & 68.4 & 46.6 & 83.8 & 75.3 & 271 & 176 & 25.4 & 99.6  & $\uparrow$ 0.3 \\
        4 & + STE       & 67.2 & 46.5 & 84.3 & 74.4 & 252 & 135 & 23.8 & 98.9  & $\downarrow$ 0.7 \\
        5 & + TTA       & 67.4 & 47.3 & 84.0 & 75.2 & 252 & 135 & 23.8 & 99.7  & $\uparrow$ 0.8 \\
        \bottomrule
    \end{tabular}
    }
    \vspace{-10pt}
\end{table}

\vspace{-8pt}
\paragraph{Component Analysis in Unified Tracking.}
\Cref{tab:unified_ablation} reports the ablation. The baseline processes 294 tokens per layer at 21.8G MACs. Mirroring the RGB case, CE, DTE, and STE (with TTA) reduces the average vision token to 188 and achieves 99.7\% performance at 16.2G MACs. TG leverages language-based priors to highlight target-relevant semantics, further raising performance to 100.0\%. These results demonstrate the robustness and generalizability of our pruning framework on unified tracking.

\begin{table}[t!]
    \caption{Ablation Study on \textbf{Unified Trackers}. $\Delta$ denotes the average performance change from the row above.} \label{tab:unified_ablation}
    \centering
    \resizebox{\columnwidth}{!}{
    \setlength{\tabcolsep}{2.0pt} 
    \begin{tabular}{c|c|ccccc|ccccc}
        \toprule
        \# & Method & 
        \rotatebox{90}{LaSOT} & 
        \rotatebox{90}{\makecell{VOT-\\RGBD22}} & 
        \rotatebox{90}{LasHeR} & 
        \rotatebox{90}{VisEvent} &
        \rotatebox{90}{OTB99} & 
         \rotatebox{90}{\makecell{Avg\\Vis Tok}} & 
         \rotatebox{90}{\makecell{Cmp\\Vis Tok}} & 
         \rotatebox{90}{\makecell{MACs\\(G)}} & 
         \rotatebox{90}{\makecell{Avg\\Perf. (\%)}} & 
         $\Delta$ \\
        \midrule
        1 & Baseline    & 73.7 & 75.5 & 59.9 & 63.3 & 68.4 & 294 & 294 & 21.8 & 100.0 & - \\
        2 & + CE        & 72.8 & 76.7 & 59.6 & 62.8 & 70.8 & 223 & 166 & 18.3 & 99.9  & $\downarrow$ 0.1 \\
        3 & + DTE       & 72.8 & 76.3 & 60.3 & 62.5 & 70.8 & 206 & 128 & 17.3 & 99.7  & $\downarrow$ 0.2 \\
        4 & + STE       & 72.3 & 76.7 & 60.1 & 61.4 & 70.1 & 188 & 90  & 16.2 & 99.3  & $\downarrow$ 0.4 \\
        5 & + TTA       & 72.4 & 75.9 & 59.4 & 61.9 & 70.7 & 188 & 90  & 16.2 & 99.7  & $\uparrow$ 0.4 \\
        6 & + TG        & 73.2 & 76.7 & 59.3 & 62.2 & 71.6 & 188 & 90  & 16.2 & 100.0 & $\uparrow$ 0.3 \\
        \bottomrule
    \end{tabular}
    }
\end{table}

\begin{figure}[t!]
    \centering
    \includegraphics[width=\columnwidth]{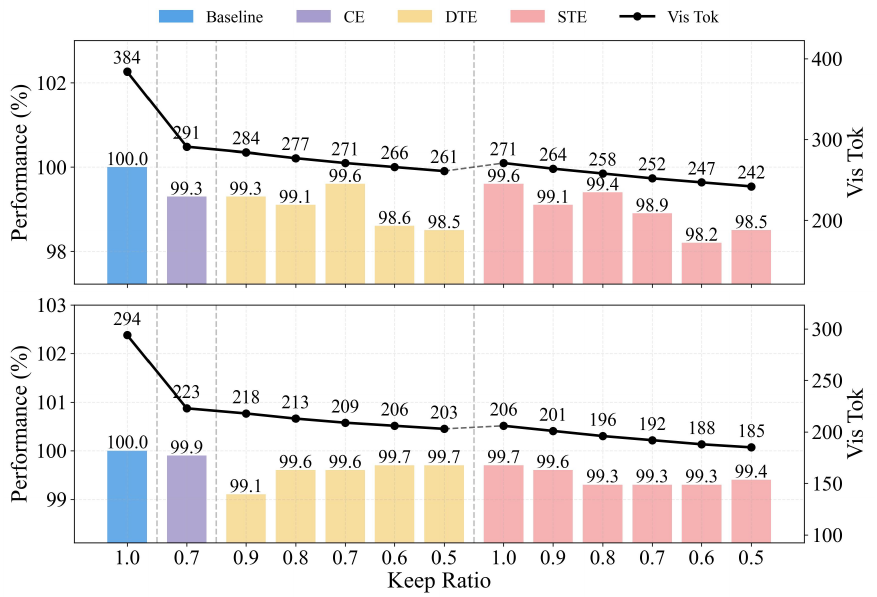}
    \caption{Ablation Study on Progressive Pruning. Performance and the number of vision tokens are reported as the keep ratio decreases and CE, DTE, and STE are progressively enabled for the RGB-based tracker (top) and unified tracker (bottom).}
    \label{fig:pruning_ratio}
    \vspace{-5pt}
\end{figure}

\vspace{-5pt}
\paragraph{Progressive Pruning Analysis.}
As illustrated in Fig.~\ref{fig:pruning_ratio}, we perform a progressive pruning study on the RGB-based and unified baselines by sequentially removing tokens from SR, DT, and ST. This schedule steadily reduces the number of visual tokens while keeping performance closely aligned with the baseline for both trackers. Across most stages, accuracy stays within 1–2\% of the original and only slightly drops under the most aggressive pruning ratio, indicating that a substantial fraction of SR, DT, and ST tokens is redundant and can be safely removed. Overall, we achieve large reductions in token count and computation with only negligible impact on tracking accuracy.

\vspace{-5pt}
\paragraph{Pruning Location Analysis.}
\label{app:dte_pos}
As shown in \cref{tab:unified_ablation_loc}, we systematically evaluate different layer-wise pruning configurations for the CE and DTE components in the unified tracker. Based on empirical results and a meticulous consideration of the trade-off between performance and efficiency, we select a configuration (\#3) that prunes CE at layers [6, 12, 18] and DTE at layers [9, 15, 21], establishing a favorable balance between performance and efficiency. A corresponding study for the RGB-based tracker is provided in \cref{app:ablation}.

\begin{table}[t!]
    \caption{Ablation Study on CTEM Location for \textbf{Unified Trackers} with a 24-layer HiViT backbone.}
    \vspace{-3pt}
    \label{tab:unified_ablation_loc}
    \centering
    \resizebox{\columnwidth}{!}{
    \setlength{\tabcolsep}{2.5pt}
    \begin{tabular}{c|cc|ccccc|cccc}
        \toprule
        \# & 
        \rotatebox{90}{\makecell{CE\\Location}} & 
        \rotatebox{90}{\makecell{DTE\\Location}} & 
        \rotatebox{90}{LaSOT} & 
        \rotatebox{90}{\makecell{VOT-\\RGBD22}} & 
        \rotatebox{90}{LasHeR} & 
        \rotatebox{90}{VisEvent} & 
        \rotatebox{90}{OTB99} & 
        \rotatebox{90}{\makecell{Avg\\Vis Tok}} & 
        \rotatebox{90}{\makecell{Cmp\\Vis Tok}} & 
        \rotatebox{90}{\makecell{MACs\\(G)}} & 
        \rotatebox{90}{\makecell{Avg\\Perf. (\%)}} \\
        \midrule
        1 & \texttt{-} & \texttt{-}      & 73.7 & 75.5 & 59.9 & 63.3 & 68.4 & 294 & 294 & 22.8 & \underline{100.0} \\
        2 & [6,12,18]  & [6,12,18]       & 71.9 & 76.0 & 59.4 & 62.5 & 70.4 & 201 & 128 & 17.0 & 99.8 \\
        3 & [6,12,18]  & [9,15,21]       & 72.8 & 76.3 & 60.3 & 62.5 & 70.8 & 206 & 128 & 17.3 & \textbf{100.6} \\
        4 & [6,12,18]  & [3,9,15]        & 71.8 & 76.1 & 60.0 & 62.5 & 70.5 & 196 & 128 & 16.7 & \underline{100.0} \\
        5 & [3,9,15]   & [6,12,18]       & 71.6 & 75.9 & 58.3 & 62.0 & 70.4 & 185 & 128 & 16.0 & 99.2 \\
        6 & [9,15,21]  & [6,12,18]       & 71.9 & 76.3 & 59.7 & 62.3 & 70.2 & 217 & 128 & 18.0 & 99.9 \\
        \bottomrule
    \end{tabular}
    }
    \vspace{-5pt}
\end{table}

\vspace{-5pt}
\paragraph{Selective Injection of Language Cues.}
To further verify the effectiveness of TG, we conduct experiments by selectively injecting language cues into different target components, including SR, DT, and ST. As shown in \cref{tab:tg_target}, we evaluate all possible injection combinations. Experimental results show that injecting language cues into DT alone (\#3) reaches 100.0\%, matching the baseline. 
These suggest that DT is the most suitable locus for language-guided pruning, which we hypothesize stems from its higher foreground concentration. In contrast, injecting language into the SR or ST leads to noticeable performance drops.

\begin{table}[t!]
    \caption{Ablation study of selectively injecting language cues into different components for text-guided token pruning. Average performance is reported relative to the baseline of unified trackers.}
    \vspace{-3pt}
    \label{tab:tg_target}
    \centering
    \resizebox{\columnwidth}{!}{
    \begin{tabular}{c|ccc|ccccc|c}
    \toprule
    \multirow{4}{*}{\#} & 
    \multicolumn{3}{c|}{Injection Target} &
    \multirow{4}{*}{\rotatebox{90}{LaSOT}} &
    \multirow{4}{*}{\rotatebox{90}{\makecell{VOT-\\RGBD22}}} &
    \multirow{4}{*}{\rotatebox{90}{LaHeR}} &
    \multirow{4}{*}{\rotatebox{90}{VisEvent}} &
    \multirow{4}{*}{\rotatebox{90}{OTB99}} &
    \multirow{4}{*}{\rotatebox{90}{\makecell{Avg\\Perf. (\%)}}} \\ [3.8ex]
    & SR & DT & ST & & & & & & \\  
    \midrule
    1 &            &            &            & 72.4 & 75.9 & 59.4 & 61.9 & 70.7 & \underline{99.7} \\
    2 & \checkmark &            &            & 72.7 & 76.0 & 59.8 & 61.7 & 70.7 & 99.4 \\
    3 &            & \checkmark &            & 73.2 & 76.7 & 59.3 & 62.2 & 71.6 & \textbf{100.0} \\
    4 &            &            & \checkmark & 72.7 & 76.3 & 59.0 & 62.2 & 70.8 & 99.3 \\
    5 & \checkmark & \checkmark &            & 72.2 & 76.5 & 59.4 & 62.0 & 70.9 & \underline{99.7} \\
    6 & \checkmark &            & \checkmark & 71.7 & 75.9 & 59.8 & 62.3 & 71.2 & 99.2 \\
    7 &            & \checkmark & \checkmark & 72.4 & 75.6 & 59.7 & 61.5 & 70.8 & \underline{99.7} \\
    8 & \checkmark & \checkmark & \checkmark & 72.9 & 76.0 & 59.3 & 62.3 & 70.5 & 99.5 \\
    \bottomrule
    \end{tabular}
    }
    \vspace{-10pt}
\end{table}

\section{Conclusion}

This work introduced \emph{UTPTrack}, a simple and unified token pruning framework that directly addresses the efficiency bottleneck in modern Transformer-based trackers. We presented the first method to jointly compress the search region, dynamic template, and static template, moving beyond the isolated pruning strategies of prior work. By leveraging attention-guided, token-type-aware mechanisms, UTPTrack effectively exploits cross-component redundancies to achieve significant computational savings. Its flexible design proves effective for both RGB tracking and more challenging multimodal and language-guided scenarios. Comprehensive evaluations on 10 benchmarks, we have shown that UTPTrack establishes a superior accuracy-efficiency balance, making it a powerful solution for developing high-performance yet practical visual trackers.

{
    \small
    \bibliographystyle{ieeenat_fullname}
    \bibliography{main}
}

\clearpage
\setcounter{page}{1}
\maketitlesupplementary
\appendix   

To complement the manuscript, this appendix presents additional content and is organized as follows:
\begin{itemize}
    \item Section~\ref{app:benchmarks}: Introduction of benchmarks.
    \item Section~\ref{app:efficiency}: Practical efficiency. 
    \item Section~\ref{app:overall_performance}: Overall performance (\cref{sec:overall_exp}).
    \item Section~\ref{app:budget}: Controlled-budget performance (\cref{sec:budget_exp}).
    \item Section~\ref{app:ablation}: Additional ablation study and analysis.
    \item Section~\ref{app:vis}: Visualization.
\end{itemize}

\vspace{-5pt}
\section{Introduction of Benchmarks}
\label{app:benchmarks}

\subsection{RGB-based Tracking Benchmarks}
\paragraph{LaSOT.}
LaSOT \cite{fan2019lasot} is a densely annotated, class-balanced large-scale long-term tracking benchmark. The benchmark consists of high-quality 1400 training set videos and 280 test set videos with an average video length of approximately 2500 frames, containing challenges of various complex scenarios. The evaluation metrics include Success (AUC) and Precision (P and P\textsubscript{Norm}), among which AUC serves as the primary metric.

\vspace{-10pt}
\paragraph{LaSOT\textsubscript{ext}.}
LaSOT\textsubscript{ext}~\cite{fan2021lasot} extends the LaSOT benchmark to include 150 videos of 15 additional new object classes, which were carefully selected outside of ImageNet. The evaluation metrics remain consistent with LaSOT.

\vspace{-10pt}
\paragraph{TrackingNet.}
TrackingNet~\cite{muller2018trackingnet} is a large-scale short-term tracking benchmark containing over 30K video sequences and over 14M bounding box annotations. The test set of TrackingNet contains 511 video sequences without publicly available ground truth.  Researchers evaluate trackers by submitting their results to the official online evaluation server, which provides AUC, P and P\textsubscript{Norm}.

\vspace{-10pt}
\paragraph{GOT-10k.}
GOT-10K~\cite{huang2019got} is a challenging, high-diversity, large-scale benchmark with more than 10K video sequences, covering most of more than 560 categories of moving objects and more than 80 categories of motion patterns. The test set has 180 video sequences, including 84 object categories and 32 motion categories. It is worth noting that GOT-10k advocates a one-shot protocol, which means that the object classes between the train set and the test set are zero-overlapped. Similar to TrackingNet, the tracking results need to be submitted to the official evaluation server to obtain the tracker's performance. The evaluation metrics include Average Overlap (AO) and Success Rates at thresholds 0.5 and 0.75 (SR\textsubscript{0.5} and SR\textsubscript{0.75}), among which AO serves as the primary metric.

\subsection{Multimodal Tracking Benchmarks}
\paragraph{VOT-RGBD22.}
VOT-RGBD22~\cite{kristan2022tenth} is a dataset designed for evaluating RGB-Depth (RGB-D) tracking algorithms, consisting of 127 video sequences. The evaluation follows the VOT protocol, including Accuracy, Robustness, and Expected Average Overlap (EAO), with EAO as the primary performance metric.

\vspace{-10pt}
\paragraph{LasHeR.}
LasHeR~\cite{li2021lasher} is a large-scale RGB-D dataset focused on long-term object tracking, containing 245 video sequences in its test set. The evaluation metrics include AUC and P, with the AUC as the primary metric.

\vspace{-10pt}
\paragraph{RGBT234.}
RGBT234~\cite{li2019rgb} is an RGB-Thermal (RGB-T) tracking dataset, comprising 234 test video sequences that cover challenging scenarios such as illumination changes, occlusions, and low visibility. The evaluation metrics include Maximum Success Rate (MSR) and Maximum Precision Rate (MPR).

\vspace{-10pt}
\paragraph{VisEvent.}
VisEvent~\cite{wang2023visevent} is a large-scale RGB-Event (RGB-E) object tracking dataset comprising 320 test video sequences. It emphasizes tracking performance under high-speed motion and low-light conditions. Researchers evaluate performance using the AUC and P metrics.

\vspace{-10pt}
\paragraph{TNL2K.}
TNL2K~\cite{wang2021towards} is a language-guided visual tracking dataset with 650 video sequences in the test set and covers a wide range of natural language descriptions. Researchers evaluate performance using AUC and P, with AUC as the primary metric.

\vspace{-10pt}
\paragraph{OTB99.}
OTB99~\cite{li2017tracking} is a classic benchmark for short-term visual object tracking, with the test set consisting of 48 video sequences. Researchers assess tracking performance based on AUC and P scores.

\section{Practical Efficiency Comparisons}
\label{app:efficiency}
As shown in Table~\ref{tab:efficiency}, UTPTrack achieves higher FPS and lower per-layer backbone latency on both GPU and CPU across resolutions for OSTrack and SUTrack, while reducing training hours via lower backbone computation.

\begin{table}[b]
\centering
\caption{Efficiency comparison. Lat.: per-layer backbone latency. GPU: NVIDIA 1080Ti; CPU: Intel Xeon Gold 6226R@2.90GHz.}
\label{tab:efficiency}
\vspace{-8pt}
\resizebox{\linewidth}{!}{
\begin{tabular}{c|ccccc}
    \toprule
    \textbf{Model} & \textbf{Train Hours} & \textbf{FPS (GPU)} & \textbf{FPS (CPU)} & \textbf{Lat. (GPU, ms)} & \textbf{Lat. (CPU, ms)} \\
    \midrule
    OSTrack\textsubscript{256} & 38.3 & 94.0  & 9.0 & 15.0 & 66.5 \\
    UTPTrack-O\textsubscript{256} & 32.6 & 95.0 & 16.7 & 12.7 & 49.5 \\
    OSTrack\textsubscript{384} & 101.4 & 39.8 & 3.2 & 36.1 & 340.8 \\
    UTPTrack-O\textsubscript{384} & 80.6 & 47.3 & 6.0 & 30.2 & 212.6 \\
    \midrule
    SUTrack\textsubscript{224} & 56.8 & 40.5 & 7.8 & 17.4 & 66.4 \\
    UTPTrack-S\textsubscript{224} & 53.3 & 42.7 & 9.3 & 17.8 & 56.4 \\
    SUTrack\textsubscript{384} & 209.2 & 27.4 & 3.4 & 63.3 & 340.2 \\
    UTPTrack-S\textsubscript{384} & 186.6 & 30.7 & 4.6 & 53.0 & 261.2 \\
    \bottomrule
\end{tabular}
}
\vspace{-8pt}
\end{table}

\begin{table*}[t!]
    \caption{SOTA comparison on RGB-based tracker across RGB-Based tracking.}
    \vspace{-5pt}
    \label{tab:rgb_trackers_ref}
    \centering
    \resizebox{\linewidth}{!}{
    \begin{tabular}{l|ccc|ccc|ccc|ccc}
        \toprule
        \multirow{2}{*}{Method} & 
        \multicolumn{3}{c|}{LaSOT} & 
        \multicolumn{3}{c|}{LaSOT\textsubscript{ext}} & 
        \multicolumn{3}{c|}{TrackingNet} & 
        \multicolumn{3}{c}{GOT-10k} \\
        \cmidrule(lr){2-4} \cmidrule(lr){5-7} \cmidrule(lr){8-10} \cmidrule(lr){11-13}
        & AUC & P\textsubscript{Norm} & P & 
          AUC & P\textsubscript{Norm} & P & 
          AUC & P\textsubscript{Norm} & P & 
          AO  & SR\textsubscript{0.5} & SR\textsubscript{0.75} \\
        \midrule
        \rowcolor{gray!20}
        \multicolumn{13}{c}{\textit{Trackers with Resolution 256}} \\
        \midrule
        OSTrack\textsubscript{256} (Baseline)  & 67.9 & 76.9 & 73.3 & 47.6 & 57.5 & 53.6 & 84.5 & 89.0 & 83.0 & 74.9 & 84.2 & 73.0 \\
        \midrule  
        OSTrack-CE\textsubscript{256}         & 67.4 & 76.5 & \underline{72.8} & \underline{46.6} & \underline{56.5} & 52.3 & \underline{83.7} & 88.3 & 82.4 & \underline{75.5} & \underline{85.5} & 73.8 \\
        OSTrack-ToMe\textsubscript{256}        & \underline{67.6} & \underline{76.6} & 72.6 & 46.5 & 56.0 & \underline{52.5} & \textbf{84.0} & 88.6 & 82.6 & 73.4 & 83.4 & 72.4 \\
        OSTrack-EViT\textsubscript{256}        & \textbf{68.4} & \textbf{77.7} & \textbf{74.1} & 46.1 & 55.7 & 51.1 & \textbf{84.0} & \underline{88.7} & \textbf{82.9} & \textbf{75.8} & \textbf{85.9} & \textbf{74.6} \\
        \textbf{UTPTrack-O\textsubscript{256}}(ours)  & 67.4 & 76.3 & 72.5 & \textbf{47.3} & \textbf{57.1} & \textbf{53.3} & \textbf{84.0} & \textbf{88.8} & \underline{82.8} & 75.2 & 85.3 & \underline{73.9} \\
        \midrule[\heavyrulewidth]   
        \rowcolor{gray!20}
        \multicolumn{13}{c}{\textit{Trackers with Higher Resolution 384}} \\
        \midrule
        OSTrack\textsubscript{384} (Baseline)      & 70.7 & 80.6 & 77.2 & 50.0 & 61.1 & 56.8 & 84.0 & 88.7 & 83.0 & 76.6 & 86.9 & 75.3 \\
        \midrule  
        OSTrack-CE\textsubscript{384}              & \underline{70.6} & \underline{80.1} & \textbf{77.0} & \underline{50.6} & \underline{61.2} & \underline{57.0} & \underline{84.1} & \underline{88.8} & 83.0 & 76.3 & 86.5 & \underline{74.9} \\
        OSTrack-ToMe\textsubscript{384}            & 70.2 & \underline{80.1} & 76.6 & 50.4 & 61.1 & 56.6 & \underline{84.1} & \underline{88.8} & \underline{83.1} & \textbf{77.1} & \textbf{87.2} & 74.7 \\
        OSTrack-EViT\textsubscript{384}            & 70.1 & 79.9 & 76.4 & \textbf{52.1} & \textbf{62.9} & \textbf{59.2} & \textbf{84.3} & \textbf{89.0} & \textbf{83.3} & 76.4 & 86.4 & \textbf{75.3} \\
        \textbf{UTPTrack-O\textsubscript{384}}(ours)      & \textbf{70.7} & \textbf{80.6} & \textbf{77.1} & 49.4 & 59.8 & 55.8 & 84.0 & 88.7 & 82.7 & \underline{76.5} & \underline{86.9} & 74.5 \\
        \bottomrule
    \end{tabular}
    }
    \vspace{-5pt}
\end{table*}

\begin{table*}[t!]
    \caption{SOTA comparison on unified tracker across RGB-Based tracking.}
    \vspace{-5pt}
    \label{tab:unified_trackers_rgb_ref}
    \centering
    \resizebox{\linewidth}{!}{
    \begin{tabular}{l|ccc|ccc|ccc|ccc}
        \toprule
        \multirow{2}{*}{Method} & 
        \multicolumn{3}{c|}{LaSOT} & 
        \multicolumn{3}{c|}{LaSOT\textsubscript{ext}} & 
        \multicolumn{3}{c|}{TrackingNet} & 
        \multicolumn{3}{c}{GOT-10k} \\
        \cmidrule(lr){2-4} \cmidrule(lr){5-7} \cmidrule(lr){8-10} \cmidrule(lr){11-13}
        & AUC & P\textsubscript{Norm} & P & 
          AUC & P\textsubscript{Norm} & P & 
          AUC & P\textsubscript{Norm} & P & 
          AO  & SR\textsubscript{0.5} & SR\textsubscript{0.75} \\
        \midrule
        \rowcolor{gray!20}
        \multicolumn{13}{c}{\textit{Trackers with Resolution 224}} \\
        \midrule
        SUTrack\textsubscript{224} (Baseline)     & 73.7 & 83.8 & 80.7 & 53.2 & 64.5 & 61.6 & 85.9 & 90.4 & 85.7 & 77.5 & 86.7 & 78.1 \\
        \midrule  
        SUTrack-CE\textsubscript{224}              & \textbf{72.8} & \textbf{82.9} & \textbf{80.2} & \underline{52.9} & \underline{63.7} & \underline{60.4} & \textbf{85.3} & \textbf{90.0} & \textbf{84.8} & \textbf{77.6} & \textbf{87.6} & \textbf{78.0} \\
        SUTrack-ToMe\textsubscript{224}            & 71.4 & 81.2 & 78.3 & 52.0 & 62.9 & 59.3 & \underline{85.2} & \textbf{90.0} & 84.6 & \underline{77.5} & \underline{87.4} & 77.5 \\
        SUTrack-EViT\textsubscript{224}            & 71.3 & 81.9 & 78.3 & 51.3 & 62.7 & 58.5 & 84.5 & \underline{89.5} & 83.5 & 76.4 & 86.5 & 75.7 \\
        SUTrack-DynamicViT\textsubscript{224}      & 57.4 & 65.5 & 59.3 & 40.8 & 49.7 & 46.6 & 65.6 & 69.1 & 60.2 & 71.6 & 81.2 & 64.7 \\
        \textbf{UTPTrack-S\textsubscript{224}}(ours)     & \underline{72.6} & \underline{82.8} & \underline{79.8} & \textbf{53.6} & \textbf{64.8} & \textbf{61.4} & \underline{85.2} & \textbf{90.0} & \underline{84.7} & 77.3 & 87.1 & \underline{77.7} \\
        \midrule[\heavyrulewidth]   
        \rowcolor{gray!20}
        \multicolumn{13}{c}{\textit{Trackers with Higher Resolution 384}} \\
        \midrule
        SUTrack\textsubscript{384} (Baseline)     & 73.9 & 83.2 & 81.2 & 52.8 & 63.3 & 60.3 & 86.8 & 90.9 & 87.3 & 79.1 & 87.6 & 80.2 \\
        \midrule  
        SUTrack-CE\textsubscript{384}               & 73.0 & 82.3 & 80.2 & \textbf{54.1} & \textbf{64.7} & \textbf{61.9} & 86.2 & 90.6 & 86.6 & \textbf{79.7} & \textbf{89.0} & \underline{80.1} \\
        SUTrack-ToMe\textsubscript{384}             & \underline{74.2} & \underline{83.8} & \textbf{81.9} & 53.0 & 63.3 & 60.5 & \underline{86.3} & \underline{90.7} & \underline{86.8} & 79.1 & 88.1 & 79.1 \\
        SUTrack-EViT\textsubscript{384}             & 73.7 & 83.4 & 81.6 & 53.0 & 63.7 & 61.0 & 86.1 & 90.6 & 86.4 & 78.4 & 87.9 & 78.1 \\
        SUTrack-DynamicViT\textsubscript{384}       & 63.0 & 72.1 & 63.0 & 44.8 & 53.4 & 47.3 & 73.5 & 80.5 & 66.3 & 69.5 & 79.0 & 61.8 \\
        \textbf{UTPTrack-S\textsubscript{384}}(ours)    & \textbf{74.3} & \textbf{83.9} & \underline{81.8} & \underline{53.6} & \underline{64.3} & \underline{61.5} & \textbf{86.4} & \textbf{90.9} & \textbf{86.9} & \underline{79.3} & \underline{88.3} & \textbf{80.3} \\
        \bottomrule
    \end{tabular}
    }
    \vspace{-5pt}
\end{table*}

\begin{table*}[t!]
    \caption{SOTA comparison on unified tracker across RGB-Depth, RGB-Thermal, RGB-Event, and RGB-Language tracking.}
    \vspace{-5pt}
    \label{tab:unified_trackers_dte_ref}
    \centering
    \resizebox{\linewidth}{!}{
    \begin{tabular}{l|ccc|cc|cc|cc|cc|cc}
        \toprule
        \multirow{2}{*}{Method} & 
        \multicolumn{3}{c|}{VOT-RGBD22} & 
        \multicolumn{2}{c|}{LasHeR} & 
        \multicolumn{2}{c|}{RGBT234} & 
        \multicolumn{2}{c|}{VisEvent} & 
        \multicolumn{2}{c|}{TNL2K} &
        \multicolumn{2}{c}{OTB99}\\
        \cmidrule(lr){2-4} \cmidrule(lr){5-6} \cmidrule(lr){7-8} \cmidrule(lr){9-10} \cmidrule(lr){11-12} \cmidrule(lr){13-14}
        & EAO & Acc & Rob & AUC & P & MSR & MPR & AUC & P & AUC & P & AUC & P \\
        \midrule
        \rowcolor{gray!20}
        \multicolumn{14}{c}{\textit{Trackers with Resolution 224}} \\
        \midrule
        SUTrack\textsubscript{224} (Baseline) & 75.5 & 82.5 & 91.3 & 59.9 & 74.8 & 70.0 & 92.1 & 63.3 & 80.7 & 67.8 & 73.8 & 68.4 & 91.1 \\
        \midrule  
        SUTrack-CE\textsubscript{224}              & \textbf{76.7} & \textbf{82.7} & \textbf{92.4} & \textbf{59.6} & \textbf{74.0} & \underline{70.1} & \underline{92.4} & \textbf{62.8} & \underline{79.9} & \underline{65.7} & \underline{71.1} & \underline{70.8} & \underline{92.7} \\
        SUTrack-ToMe\textsubscript{224}            & 76.1 & \textbf{82.7} & 91.8 & \underline{59.2} & 73.7 & 69.4 & 91.8 & 61.7 & 78.9 & \textbf{65.9} & \textbf{71.4} & 70.5 & 91.6 \\
        SUTrack-EViT\textsubscript{224}            & 75.0 & 82.1 & 90.9 & 58.6 & 73.0 & 69.2 & 91.6 & 62.0 & 79.4 & 65.0 & 70.0 & \underline{70.8} & 91.9 \\
        SUTrack-DynamicViT\textsubscript{224}      & 62.8 & 74.2 & 84.4 & 46.1 & 57.5 & 63.2 & 85.1 & 50.9 & 71.8 & 55.6 & 55.0 & 65.8 & 86.8 \\
        \textbf{UTPTrack-S\textsubscript{224}}(ours)     & \underline{76.4} & \textbf{82.7} & \underline{92.1} & \underline{59.2} & \underline{73.9} & \textbf{70.4} & \textbf{92.9} & \underline{62.6} & \textbf{80.0} & 65.6 & 71.0 & \textbf{71.9} & \textbf{93.7} \\
        \midrule[\heavyrulewidth]   
        \rowcolor{gray!20}
        \multicolumn{14}{c}{\textit{Trackers with Higher Resolution 384}} \\
        \midrule
        SUTrack\textsubscript{384} (Baseline) & 76.0 & 83.2 & 91.5 & 60.6 & 75.3 & 69.2 & 92.4 & 63.0 & 79.3 & 68.2 & 74.7 & 69.2 & 91.0 \\
        \midrule  
        SUTrack-CE\textsubscript{384}              & \textbf{77.9} & \underline{83.6} & \textbf{92.8} & \textbf{60.4} & \textbf{75.0} & \textbf{69.7} & \underline{92.0} & \textbf{62.9} & \underline{79.5} & 66.2 & 72.6 & 70.0 & 90.6 \\
        SUTrack-ToMe\textsubscript{384}            & \underline{76.8} & \textbf{83.8} & 91.5 & \underline{59.7} & 74.1 & \underline{69.4} & \underline{92.0} & 62.6 & 79.3 & \underline{66.4} & \underline{72.8} & 70.8 & 91.8 \\
        SUTrack-EViT\textsubscript{384}            & 77.1 & 83.2 & \underline{92.5} & 58.8 & 73.2 & 68.2 & 90.8 & 62.1 & 79.2 & 66.2 & 72.7 & \textbf{72.5} & \textbf{94.6} \\
        SUTrack-DynamicViT\textsubscript{384}      & 71.2 & 81.0 & 87.8 & 51.6 & 63.5 & 56.7 & 82.2 & 56.4 & 73.4 & 58.9 & 58.0 & 68.0 & 86.5 \\
        \textbf{UTPTrack-S\textsubscript{384}}(ours)     & \underline{76.8} & \textbf{83.8} & \underline{91.6} & \textbf{60.4} & \underline{74.8} & \textbf{69.7} & \textbf{92.2} & \underline{62.8} & \textbf{79.7} & \textbf{66.6} & \textbf{72.9} & \underline{72.2} & \underline{93.7} \\
        \bottomrule
    \end{tabular}
    }
    \vspace{-5pt}
\end{table*}

\section{Detailed Overall Performance Comparisons}
\label{app:overall_performance}
In \cref{sec:overall_exp}, we validate the effectiveness of UTPTrack under two tracking paradigms: RGB-based tracking and Unified tracking, which includes RGB-Depth (RGB-D), RGB-Thermal (RGB-T), RGB-Event (RGB-E), and RGB-Language (RGB-Lang) tasks. Using OSTrack and SUTrack as representative base models, we compare UTPTrack against state-of-the-art token compression methods, including CE, ToMe, EViT, and DynamicViT, across varying input resolutions. We report comprehensive results for Overall Performance, with detailed tabular data and analysis provided in this section. 

\vspace{-6pt}
\paragraph{RGB-based Tracking.}
We evaluate UTPTrack on four large-scale RGB-based tracking benchmarks, covering both long-term (LaSOT, LaSOT\textsubscript{ext}) and short-term (TrackingNet, GOT-10k) scenarios. The results are summarized in Tab.~\ref{tab:rgb_trackers_ref} and Tab.~\ref{tab:unified_trackers_rgb_ref}. 
For the RGB-based variant, UTPTrack-O maintains the accuracy of base model across all four benchmarks and both resolutions. At 256 resolution, UTPTrack-O\textsubscript{256} closely matches base model on LaSOT, LaSOT\textsubscript{ext}, and TrackingNet while slightly improving AO on GOT-10k, and remains competitive with other compression methods. At 384 resolution, UTPTrack-O\textsubscript{384} similarly matches the performance of base model across all benchmarks.
In contrast, the unified variant, UTPTrack-S, consistently delivers strong results across all benchmarks. UTPTrack-S\textsubscript{224} improves the AUC on LaSOT\textsubscript{ext} and TrackingNet over base model, while UTPTrack-S\textsubscript{384} achieves the best or second-best scores on all four datasets. These results demonstrate that our pruning strategy effectively preserves accuracy across different resolutions and tracking paradigms.

\vspace{-5pt}
\paragraph{RGB-D/T/E Tracking.}

As shown in Tab.\ref{tab:unified_trackers_dte_ref}, UTPTrack-S\textsubscript{224} and UTPTrack-S\textsubscript{384} consistently rank among the top two across all benchmarks, including VOT-RGBD22 (RGB-D), LasHeR and RGBT234 (RGB-T), and VisEvent (RGB-E). This demonstrates the robustness and generalizability of our pruning strategy across diverse modalities and dataset scales, while maintaining strong performance and high efficiency.

\vspace{-5pt}
\paragraph{RGB-Lang Tracking.}


As shown in ~\cref{tab:unified_trackers_dte_ref}, UTPTrack-S significantly outperforms the baseline model on the short-term tracking dataset OTB99 across different resolutions. It achieves gains of 3.5\% at 224 resolution and 2.0\% at 384 resolution. On TNL2K, however, the performance of UTPTrack-S decreases relative to the baseline model, likely due to compression-induced performance loss. Even so, UTPTrack-S still delivers competitive performance on TNL2K, and at 384 resolution it achieves the best results among all compared methods. Such degradation across all compressed models may be attributed to the limited precision of the initial language descriptions in later stages of long videos, especially in long-term sequences, which reduces the reliability of language-guided cues.

\section{Detaild Controlled-Budget Performance Comparisons}
\label{app:budget}

In ~\cref{sec:budget_exp}, we conduct controlled-budget experiments under fixed compression ratios to ensure fair comparisons for RGB-based and unified tracking, with comprehensive results presented in this section. 

\vspace{-5pt}
\paragraph{RGB-based Tracking.}
For RGB-based tracking, we compare UTPTrack with other methods on four widely used RGB-based benchmarks. As shown in \cref{tab:rgb_trackers_budget_ref}, UTPTrack-O consistently outperforms all other methods across all pruning rates. Notably, with 18.8\% of the visual tokens pruned, UTPTrack-O\textsubscript{256} outperforms the baseline model by 0.2\%, suggesting that removing redundant tokens can even lead to performance gains. Similarly, in \cref{tab:unified_trackers_rgb_budget_ref}, UTPTrack-S surpasses all other compression methods across all pruning rates, and at a retained ratio of 35.4\% visual tokens, it achieves a 1.0\% improvement over the second-best method.

\begin{table*}[t!]
    \caption{Performance comparisons under different vision token compression settings for \textbf{RGB-based tracking}. The baseline uses 384 tokens. The first row shows raw scores, and the second row reports percentages relative to the upper bound.}
    \vspace{-5pt}
    \label{tab:rgb_trackers_budget_ref}
    \centering
    \resizebox{\linewidth}{!}{
    \setlength{\tabcolsep}{2pt} 
    \begin{tabular}{l|ccc|ccc|ccc|ccc|c}
        \toprule
        \multirow{2}{*}{Method} & 
        \multicolumn{3}{c|}{LaSOT} & 
        \multicolumn{3}{c|}{LaSOT\textsubscript{ext}} & 
        \multicolumn{3}{c|}{TrackingNet} & 
        \multicolumn{3}{c|}{GOT-10k} & 
        \multirow{2}{*}{Avg Perf.(\%)} \\
        \cmidrule(lr){2-4} \cmidrule(lr){5-7} \cmidrule(lr){8-10} \cmidrule(lr){11-13}
        & AUC & P\textsubscript{Norm} & P & 
        AUC & P\textsubscript{Norm} & P & 
        AUC & P\textsubscript{Norm} & P & 
        AO  & SR\textsubscript{0.5} & SR\textsubscript{0.75} 
        & \\
        \midrule
        \multicolumn{14}{c}{\textit{Upper Bound, 384 Tokens (100\%)}} \\
        \midrule
        \multirow{2}{*}{OSTrack$_{256}$ (Baseline)} & 67.9 & 76.9 & 73.3 & 47.6 & 57.5 & 53.6 & 84.5 & 89.0 & 83.0 & 74.9 & 84.2 & 73.0 & \multirow{2}{*}{100\%} \\
                                                               &100\% &100\% &100\% &100\% &100\% &100\% &100\% &100\% &100\% &100\% &100\% &100\% & \\
        \midrule
        \multicolumn{14}{c}{\textit{Retain 87.2\% Tokens ($\approx$ 335 tokens, $\downarrow$ 18.8\%)}} \\
        \midrule
        \multirow{2}{*}{OSTrack-CE\textsubscript{256}} & 67.1 & 75.8 & 72.0 & 46.9 & 56.7 & 52.4 & 84.1 & 88.8 & 83.0 & 74.9 & 85.0 & 73.5 & \multirow{2}{*}{\underline{99.3\%}($\downarrow0.7\%$)} \\
        & 98.9\% & 98.6\% & 98.3\% & 98.6\% & 98.6\% & 97.7\% & 99.6\% & 99.8\% & 99.9\% & 100.0\% & 101.0\% & 100.7\% & \\
        
        \cmidrule(lr){1-14}
        \multirow{2}{*}{OSTrack-ToMe\textsubscript{256}} & 68.0 & 76.9 & 72.9 & 46.4 & 56.3 & 51.5 & 83.8 & 88.7 & 82.7 & 74.3 & 84.2 & 73.0 & \multirow{2}{*}{99.0\%($\downarrow1.0\%$)} \\
                                                         & 100.2\% & 99.9\% & 99.5\% & 97.5\% & 98.0\% & 96.1\% & 99.3\% & 99.6\% & 99.6\% & 99.2\% & 100.0\% & 100.0\% & \\
        \cmidrule(lr){1-14}
        \multirow{2}{*}{OSTrack-EViT\textsubscript{256}} & 66.8 & 75.7 & 71.7 & 46.7 & 56.3 & 52.3 & 84.1 & 88.9 & 82.8 & 75.0 & 85.2 & 73.9 & \multirow{2}{*}{99.0\%($\downarrow1.0\%$)} \\
                                                         & 98.4\% & 98.4\% & 97.9\% & 98.1\% & 98.1\% & 97.5\% & 99.5\% & 99.8\% & 99.7\% & 100.1\% & 101.2\% & 101.2\% & \\
        \cmidrule(lr){1-14}
        \multirow{2}{*}{UTPTrack-O\textsubscript{256}} & 67.1 & 75.7 & 72.2 & 48.3 & 58.3 & 54.4 & 84.0 & 88.7 & 83.0 & 75.6 & 85.8 & 74.7 & \multirow{2}{*}{\textbf{100.2\%}($\uparrow0.2\%$)} \\
                                                     & 98.9\% & 98.5\% & 98.5\% & 101.5\% & 101.5\% & 101.4\% & 99.5\% & 99.6\% & 99.9\% & 100.9\% & 101.9\% & 102.3\% & \\
        \midrule
        \multicolumn{14}{c}{\textit{Retain 75.5\% Tokens ($\approx$ 290 tokens, $\downarrow$ 24.5\%)}} \\
        \midrule
        \multirow{2}{*}{OSTrack-CE\textsubscript{256}} & 67.3 & 76.2 & 72.6 & 47.3 & 57.2 & 53.1 & 84.2 & 89.0 & 83.2 & 74.2 & 84.1 & 72.8 & \multirow{2}{*}{99.3\%($\downarrow0.7\%$)} \\
                                                       & 99.1\% & 99.1\% & 99.1\% & 99.4\% & 99.5\% & 99.0\% & 99.7\% & 100.0\% & 100.2\% & 99.1\% & 99.9\% & 99.7\% & \\
        \cmidrule(lr){1-14}
        \multirow{2}{*}{OSTrack-ToMe\textsubscript{256}} & 68.2 & 77.1 & 73.2 & 46.5 & 56.0 & 52.2 & 84.2 & 89.0 & 83.2 & 74.9 & 85.1 & 73.6 & \multirow{2}{*}{\underline{99.4\%}($\downarrow0.6\%$)} \\
                                                         & 100.4\% & 100.2\% & 100.0\% & 97.6\% & 97.5\% & 97.4\% & 99.7\% & 100.0\% & 100.1\% & 100.0\% & 101.1\% & 100.8\% & \\
        \cmidrule(lr){1-14}
        \multirow{2}{*}{OSTrack-EViT\textsubscript{256}} & 67.8 & 76.8 & 73.4 & 47.1 & 56.6 & 52.8 & 83.8 & 88.5 & 82.5 & 72.9 & 82.8 & 70.4 & \multirow{2}{*}{98.8\%($\downarrow1.2\%$)} \\
                                                         & 99.8\% & 99.9\% & 100.2\% & 99.0\% & 98.6\% & 98.4\% & 99.2\% & 99.4\% & 99.4\% & 97.3\% & 98.3\% & 96.4\% & \\
        \cmidrule(lr){1-14}
        \multirow{2}{*}{UTPTrack-O\textsubscript{256}} & 67.6 & 76.3 & 72.9 & 47.3 & 57.2 & 53.0 & 83.8 & 88.4 & 82.7 & 74.8 & 85.1 & 73.5 & \multirow{2}{*}{\textbf{99.5\%}($\downarrow0.5\%$)} \\
                                                     & 99.5\% & 99.3\% & 99.5\% & 99.5\% & 99.6\% & 98.9\% & 99.2\% & 99.3\% & 99.6\% & 99.9\% & 101.1\% & 100.7\% & \\
        \midrule
        \multicolumn{14}{c}{\textit{Retain 65.6\% Tokens ($\approx$ 252 tokens, $\downarrow$ 34.4\%)}} \\
        \midrule
        \multirow{2}{*}{OSTrack-CE\textsubscript{256}} & 67.2 & 76.1 & 72.5 & 46.5 & 56.3 & 51.8 & 83.3 & 88.2 & 82.0 & 72.8 & 82.9 & 70.9 & \multirow{2}{*}{\underline{98.1\%}($\downarrow1.9\%$)} \\
                                                       & 99.0\% & 99.0\% & 98.9\% & 97.8\% & 97.9\% & 96.6\% & 98.6\% & 99.0\% & 98.7\% & 97.2\% & 98.5\% & 97.1\% & \\
        \cmidrule(lr){1-14}
        \multirow{2}{*}{OSTrack-ToMe\textsubscript{256}} & 66.3 & 75.7 & 71.1 & 46.0 & 55.9 & 50.9 & 82.5 & 87.5 & 80.2 & 71.0 & 81.3 & 67.9 & \multirow{2}{*}{96.7\%($\downarrow3.3\%$)} \\
                                                         & 97.7\% & 98.4\% & 97.0\% & 96.7\% & 97.2\% & 94.9\% & 97.7\% & 98.2\% & 96.6\% & 94.8\% & 96.6\% & 93.0\% & \\
        \cmidrule(lr){1-14}
        \multirow{2}{*}{OSTrack-EViT\textsubscript{256}} & 68.1 & 77.2 & 73.2 & 46.1 & 55.8 & 51.2 & 83.0 & 87.9 & 81.7 & 72.8 & 83.3 & 69.7 & \multirow{2}{*}{\underline{98.1\%}($\downarrow1.9\%$)} \\
                                                         & 100.3\% & 100.4\% & 99.9\% & 96.8\% & 97.1\% & 95.5\% & 98.2\% & 98.7\% & 98.4\% & 97.2\% & 98.9\% & 95.5\% & \\
        \cmidrule(lr){1-14}
        \multirow{2}{*}{UTPTrack-O\textsubscript{256}} & 68.2 & 77.3 & 74.0 & 46.2 & 56.1 & 51.8 & 84.0 & 88.9 & 83.1 & 74.9 & 85.0 & 73.3 & \multirow{2}{*}{\textbf{99.2\%}($\downarrow0.8\%$)} \\
                                                     & 100.4\% & 100.5\% & 101.1\% & 97.0\% & 97.7\% & 96.6\% & 99.5\% & 99.8\% & 100.1\% & 100.0\% & 101.0\% & 100.4\% & \\
        \bottomrule
    \end{tabular}
    }
    \vspace{-5pt}
\end{table*}

\vspace{-5pt}
\paragraph{RGB-D/T/E Tracking.}
As shown in \cref{tab:unified_trackers_dte_budget_ref}, UTPTrack outperforms the baseline on both the RGB-D and RGB-T benchmarks across different pruning ratios. On the VOT-RGBD22 dataset, it surpasses the baseline by 1.3\% when pruning 25.6\% visual tokens. On the RGB-E benchmark, UTPTrack consistently ranks among the top two methods. These results demonstrate that UTPTrack generalizes effectively across diverse multimodal tracking scenarios.

\vspace{-5pt}
\paragraph{RGB-Lang Tracking.}
As shown in \cref{tab:unified_trackers_dte_budget_ref}, UTPTrack maintains competitive performance across different compression rates on the RGB-Language benchmark. At all compression rates, UTPTrack outperforms the baseline on the short-term tracking dataset OTB99. With 25.6\% visual tokens pruned, UTPTrack surpasses the baseline by 3.3\%, and even at a pruning rate of 64.6\%, it still leads the baseline by 2.7\%. On TNL2K, UTPTrack consistently ranks in the top two compared to other compression methods.

\section{Additional Ablation Study}
\label{app:ablation}

\subsection{The Effect of Pruning Location for RGB-based Tracking}


To analyze the impact of pruning locations, we systematically vary the pruning positions of both the CE and DTE modules across different transformer layers, with detailed results summarized in \cref{tab:rgb_ablation_loc}.
Based on empirical results and a meticulous consideration of the performance-efficiency trade-off, we select the pruning configuration (\#3) that prunes CE at layers [3, 6, 9], and DTE at layers [4, 7, 10]. This arrangement achieves an optimal balance, maintaining competitive tracking performance while substantially reducing the computational overhead.

\begin{table}[htbp]
    \caption{Ablation Study on CTEM Location for \textbf{RGB-based Trackers} with a 12-layer ViT backbone.}
    \label{tab:rgb_ablation_loc}
    \centering
    \resizebox{\columnwidth}{!}{
    \setlength{\tabcolsep}{4.0pt}
    \begin{tabular}{c|cc|cccc|cccc}
        \toprule
        \# & 
        \rotatebox{90}{\makecell{CE\\Location}} & 
        \rotatebox{90}{\makecell{DTE\\Location}} & 
        \rotatebox{90}{LaSOT} & 
        \rotatebox{90}{LaSOT\textsubscript{ext}} & 
        \rotatebox{90}{TrackingNet} & 
        \rotatebox{90}{GOT-10k} & 
        \rotatebox{90}{\makecell{Avg\\Vis Tok}} & 
        \rotatebox{90}{\makecell{Cm\\Vis Tok}} & 
        \rotatebox{90}{\makecell{MACs\\(G)}} & 
        \rotatebox{90}{\makecell{Avg\\Perf. (\%)}} \\
        \midrule
        1 & -- & -- & 67.9 & 47.6 & 84.5 & 74.9 & 384.0 & 384 & 34.5 & 100.0 \\
        2 & [3,6,9]  & [3,6,9]  & 67.6 & 47.0 & 83.8 & 74.5 & 267.8 & 176 & 25.1 & 99.3 \\
        3 & [3,6,9]  & [4,7,10] & 68.4 & 46.6 & 83.8 & 75.3 & 271.2 & 176 & 25.4 & \textbf{99.6} \\
        4 & [3,6,9]  & [2,5,8]  & 67.4 & 45.7 & 83.8 & 75.4 & 264.3 & 176 & 24.8 & 98.8 \\
        5 & [2,5,8]  & [3,6,9]  & 68.0 & 46.4 & 83.9 & 75.7 & 253.8 & 176 & 23.9 & \underline{99.5} \\
        6 & [4,7,10] & [3,6,9]  & 67.5 & 46.5 & 83.9 & 73.4 & 281.7 & 176 & 26.3 & 98.6 \\
        \bottomrule
    \end{tabular}
    }
\end{table}


\subsection{The Effect of Spatial Priors in Token Type-Aware Strategy}
TTA does not replace attention-based importance estimation, but stabilizes static template pruning, which is sensitive due to limited tokens and foreground–background imbalance. The bounding-box prior uses only initialization information. As shown in Table~\ref{tab:prior}, attention-based pruning without priors consistently outperforms random pruning, while adding spatial priors yields modest but consistent gains. These results indicate that TTA complements attention-based pruning by improving stability under high compression ratios.

\begin{table}[htbp]
\caption{Effect of spatial priors in attention-guided pruning.}
\label{tab:prior}
\centering
\resizebox{\columnwidth}{!}{
\setlength{\tabcolsep}{4.0pt} 
\begin{tabular}{c|cccc}
    \toprule
    \textbf{Method} & \textbf{LaSOT} & \textbf{LaSOT\textsubscript{ext}} & \textbf{TrackingNet} & \textbf{GOT-10k} \\
    \midrule
    Attn-based Pruning w/o Priors & 67.2 & 46.5 & \textbf{84.3} & 74.4 \\
    Attn-based Pruning w/ Priors  & \textbf{67.4} & \textbf{47.3} & 84.0 & \textbf{75.2} \\
    Random Pruning w/ Priors      & 67.0 & 46.8 & 83.9 & 74.2 \\
    \bottomrule
\end{tabular}
}
\end{table}

\subsection{The Effect of Token Type-Aware Strategy for Unified Tracking}


\Cref{tab:unified_ablation_tta_loc} compares the effects of three foreground bonus strategies on ST pruning for unified trackers, corresponding to the Full, Soft and All variants introduced in~\cref{sec:tta}. Among them, the Soft bonus delivers the best overall performance across the five benchmarks, achieving an average accuracy of 99.8\%, outperforming the Full and All strategies by approximately 0.5\% and 0.7\%, respectively. Under a strict pruning ratio, the Full bonus rewards only patches that lie entirely within the target box, making it overly conservative and prone to underestimating foreground tokens near object boundaries. Conversely, the All bonus assigns the maximum reward as long as a patch merely intersects the box, which is excessively permissive and easily introduces background noise. The Soft bonus, computed as the mean mask value within each patch, provides a more fine-grained estimate of foreground coverage and yields smoother boundary transitions, making it a more reliable continuous signal for guiding pruning.


\begin{table}[htbp]
    \caption{Ablation Study on bonus for Unified Trackers.}
    \label{tab:unified_ablation_tta_loc}
    \centering
    \resizebox{\columnwidth}{!}{
    \setlength{\tabcolsep}{2.5pt}
    \begin{tabular}{c|c|ccccc|c}
        \toprule
        \# & 
        Bonus & 
        LaSOT & 
        \makecell{VOT-\\RGBD22} & 
        LasHeR & 
        VisEvent & 
        OTB99 & 
        \makecell{Avg\\Perf. (\%)} \\
        \midrule
        1 & Full    & 71.7 & 75.7 & 59.5 & 61.7 & 69.7 & 99.3 \\
        2 & Soft    & 72.4 & 75.9 & 59.4 & 61.9 & 70.7 & \textbf{99.8} \\
        3 & All     & 72.2 & 75.6 & 59.5 & 61.2 & 69.4 & 99.1 \\
        \bottomrule
    \end{tabular}
    }
\end{table}

\begin{table*}[t!]
    \caption{Performance comparisons under different vision token compression configurations across \textbf{unified tracking} on RGB-based tracking benchmark. The vanilla number of visual tokens is 294. The first line of each method is the raw performance of benchmarks, and the second line is the proportion relative to the upper limit. The average performance listed is calculated across all 10 benchmarks.}
    \vspace{-5pt}
    \label{tab:unified_trackers_rgb_budget_ref}
    \centering
    \resizebox{\linewidth}{!}{
    \setlength{\tabcolsep}{4pt} 
    \begin{tabular}{l|ccc|ccc|ccc|ccc|c}
        \toprule
        \multirow{2}{*}{Method} & 
        \multicolumn{3}{c|}{LaSOT} & 
        \multicolumn{3}{c|}{LaSOT\textsubscript{ext}} & 
        \multicolumn{3}{c|}{TrackingNet} & 
        \multicolumn{3}{c|}{GOT-10k} & 
        \multirow{2}{*}{\makecell{Avg\\Perf.(\%)}} \\
        \cmidrule(lr){2-4} \cmidrule(lr){5-7} \cmidrule(lr){8-10} \cmidrule(lr){11-13}
        & AUC & P\textsubscript{Norm} & P & 
        AUC & P\textsubscript{Norm} & P & 
        AUC & P\textsubscript{Norm} & P & 
        AO  & SR\textsubscript{0.5} & SR\textsubscript{0.75} 
        & \\
        \midrule
        \multicolumn{14}{c}{\textit{Upper Bound, 294 Tokens (100\%)}} \\
        \midrule
        \multirow{2}{*}{SUTrack\textsubscript{224} (Baseline)} 
        & 73.7 & 83.8 & 80.7 & 53.2 & 64.5 & 61.6 & 85.9 & 90.4 & 85.7 & 77.5 & 86.7 & 78.1 & \multirow{2}{*}{100\%} \\
        & 100\% &100\% &100\% &100\% &100\% &100\% &100\% &100\% &100\% &100\% &100\% &100\% & \\
        \midrule
        \multicolumn{14}{c}{\textit{Retain 71.4\% Tokens ($\approx$ 218 tokens, $\downarrow$ 25.6\%)}} \\
        \midrule
        \multirow{2}{*}{SUTrack-CE\textsubscript{224}} 
        & 72.4 & 82.4 & 79.2 & 52.8 & 63.6 & 60.2 & 85.1 & 89.7 & 84.4 & 77.4 & 87.0 & 78.0 & \multirow{2}{*}{\underline{99.5}\%($\downarrow0.5\%$)} \\
        & 98.3\% &98.4\% &98.1\% &99.1\% &98.7\% &97.7\% &99.0\% &99.2\% &98.5\% &99.9\% &100.3\% &99.9\% \\
        \cmidrule(lr){1-14}
        \multirow{2}{*}{SUTrack-ToMe\textsubscript{224}} 
        & 72.1 & 82.0 & 78.9 & 52.6 & 63.7 & 59.7 & 85.4 & 90.0 & 84.7 & 76.9 & 86.5 & 77.2 & \multirow{2}{*}{99.4\%($\downarrow0.6\%$)} \\
        & 97.8\% &97.8\% &97.7\% &98.8\% &98.8\% &96.9\% &99.4\% &99.6\% &98.9\% &99.2\% &99.8\% &98.8\% & \\
        \cmidrule(lr){1-14}
        \multirow{2}{*}{SUTrack-EViT\textsubscript{224}} 
        & 71.6 & 81.9 & 79.0 & 52.3 & 63.2 & 59.5 & 84.9 & 89.7 & 84.1 & 76.9 & 87.2 & 76.8 & \multirow{2}{*}{99.0\%($\downarrow1.0\%$)} \\
        & 97.1\% &97.7\% &97.8\% &98.3\% &98.1\% &96.6\% &98.8\% &99.2\% &98.2\% &99.2\% &100.6\% &98.3\% & \\
        \cmidrule(lr){1-14}
        \multirow{2}{*}{SUTrack-DyViT\textsubscript{224}} 
        & 68.8 & 81.6 & 71.1 & 49.7 & 63.0 & 55.3 & 80.2 & 86.4 & 74.9 & 74.6 & 87.1 & 72.1 & \multirow{2}{*}{96.5\%($\downarrow3.5\%$)} \\
        & 93.4\% &97.4\% &88.0\% &93.4\% &97.8\% &89.8\% &93.4\% &95.6\% &87.4\% &96.3\% &100.5\% &92.3\% & \\
        \cmidrule(lr){1-14}
        \multirow{2}{*}{UTPTrack-S\textsubscript{224}} 
        & 72.1 & 82.2 & 78.9 & 52.9 & 63.8 & 60.1 & 85.1 & 89.7 & 84.5 & 77.7 & 87.4 & 78.2 & \multirow{2}{*}{\textbf{99.8\%($\downarrow0.2\%$)}} \\
        & 97.9\% &98.1\% &97.7\% &99.4\% &98.9\% &97.5\% &99.0\% &99.3\% &98.6\% &100.3\% &100.8\% &100.1\% & \\
        \midrule
        \multicolumn{14}{c}{\textit{Retain 52.0\% Tokens ($\approx$ 153 tokens, $\downarrow$ 48.0\%)}} \\
        \midrule
        \multirow{2}{*}{SUTrack-CE\textsubscript{224}} 
        & 72.0 & 81.8 & 78.9 & 52.8 & 63.9 & 60.4 & 85.3 & 90.0 & 84.7 & 77.1 & 86.8 & 77.7 & \multirow{2}{*}{\underline{99.2}\%($\downarrow0.8\%$)} \\
        & 97.7\% &97.6\% &97.7\% &99.2\% &99.1\% &97.9\% &99.3\% &99.6\% &98.8\% &99.5\% &100.1\% &99.5\% & \\
        \cmidrule(lr){1-14}
        \multirow{2}{*}{SUTrack-ToMe\textsubscript{224}} 
        & 71.4 & 81.2 & 78.3 & 52.0 & 62.9 & 59.3 & 85.2 & 90.0 & 84.6 & 77.5 & 87.4 & 77.5 & \multirow{2}{*}{99.0\%($\downarrow1.0\%$)} \\
        & 96.9\% &96.9\% &96.9\% &97.8\% &97.6\% &96.2\% &99.1\% &99.6\% &98.7\% &100.0\% &100.8\% &99.2\% & \\
        \cmidrule(lr){1-14}
        \multirow{2}{*}{SUTrack-EViT\textsubscript{224}} 
        & 71.2 & 81.6 & 78.4 & 51.9 & 61.1 & 58.9 & 84.7 & 89.7 & 84.0 & 76.9 & 87.2 & 76.9 & \multirow{2}{*}{98.5\%($\downarrow1.5\%$)} \\
        & 96.6\% &97.4\% &97.1\% &97.5\% &94.8\% &95.6\% &98.6\% &99.3\% &98.0\% &99.2\% &100.6\% &98.5\% & \\
        \cmidrule(lr){1-14}
        \multirow{2}{*}{SUTrack-DyViT\textsubscript{224}} 
        & 53.2 & 60.6 & 55.3 & 36.5 & 44.1 & 41.7 & 58.2 & 62.4 & 52.2 & 52.2 & 60.0 & 34.8 & \multirow{2}{*}{78.8\%($\downarrow21.2\%$)} \\
        & 72.1\% &72.3\% &68.5\% &68.5\% &68.4\% &67.6\% &67.7\% &69.0\% &61.0\% &67.4\% &69.2\% &44.6\% & \\
        \cmidrule(lr){1-14}
        \multirow{2}{*}{UTPTrack-S\textsubscript{224}} 
        & 72.5 & 82.5 & 79.6 & 53.1 & 64.0 & 60.5 & 85.5 & 90.1 & 84.9 & 77.1 & 86.8 & 77.5 & \multirow{2}{*}{\textbf{99.5}\%($\downarrow0.5\%$)} \\
        & 98.4\% &98.5\% &98.6\% &99.7\% &99.3\% &98.2\% &99.5\% &99.7\% &99.1\% &99.5\% &100.1\% &99.2\% & \\
        \midrule
        \multicolumn{14}{c}{\textit{Retain 35.4\% Tokens ($\approx$ 104 tokens, $\downarrow$ 64.6\%)}} \\
        \midrule
        \multirow{2}{*}{SUTrack-CE\textsubscript{224}} 
        & 70.5 & 80.5 & 77.2 & 51.8 & 63.0 & 59.3 & 84.8 & 89.5 & 83.9 & 75.9 & 85.9 & 75.3 & \multirow{2}{*}{\underline{98.3}\%($\downarrow1.7\%$)} \\
        & 95.7\% &96.1\% &95.6\% &97.3\% &97.7\% &96.2\% &98.7\% &99.1\% &97.9\% &97.9\% &99.1\% &96.4\% & \\
        \cmidrule(lr){1-14}
        \multirow{2}{*}{SUTrack-ToMe\textsubscript{224}} 
        & 67.9 & 73.9 & 70.7 & 48.6 & 55.9 & 53.8 & 80.0 & 83.1 & 75.5 & 71.6 & 81.2 & 64.7 & \multirow{2}{*}{92.5\%($\downarrow7.5\%$)} \\
        & 92.1\% &88.1\% &87.6\% &91.3\% &86.6\% &87.2\% &93.1\% &92.0\% &88.1\% &92.4\% &93.7\% &82.8\% & \\
        \cmidrule(lr){1-14}
        \multirow{2}{*}{SUTrack-EViT\textsubscript{224}} 
        & 69.6 & 80.0 & 76.5 & 50.6 & 61.8 & 57.6 & 84.3 & 89.2 & 83.2 & 74.0 & 84.3 & 72.4 & \multirow{2}{*}{96.7\%($\downarrow3.3\%$)} \\
        & 94.5\% &95.4\% &94.7\% &95.1\% &95.9\% &93.4\% &98.1\% &98.8\% &97.0\% &95.5\% &97.2\% &92.7\% & \\
        \cmidrule(lr){1-14}
        \multirow{2}{*}{SUTrack-DyViT\textsubscript{224}} 
        & 4.4 & 3.6 & 3.6 & 2.9 & 5.8 & 2.5 & 8.8 & 6.7 & 5.7 & 7.4 & 7.1 & 0.8 & \multirow{2}{*}{14.7\%($\downarrow85.3\%$)} \\
        & 5.9\% &4.2\% &4.4\% &5.4\% &9.0\% &4.1\% &10.3\% &7.4\% &6.7\% &9.5\% &8.2\% &1.0\% & \\
        \cmidrule(lr){1-14}
        \multirow{2}{*}{UTPTrack-S\textsubscript{224}} 
        & 72.3 & 82.3 & 79.3 & 52.7 & 63.7 & 60.0 & 85.2 & 89.7 & 84.4 & 77.5 & 87.1 & 77.9 & \multirow{2}{*}{\textbf{99.3}\%($\downarrow0.7\%$)} \\
        & 98.2\% &98.2\% &98.2\% &99.0\% &98.8\% &97.4\% &99.1\% &99.3\% &98.5\% &100.0\% &100.5\% &99.7\% & \\
        \bottomrule
    \end{tabular}
    }
    \vspace{-5pt}
\end{table*}

\begin{table*}[t!]
    \caption{Performance comparisons under different vision token compression configurations across \textbf{unified tracking} on RGB-D/T/E/Lang tracking benchmark. The vanilla number of visual tokens is 294. The first line of each method is the raw performance of benchmarks, and the second line is the proportion relative to the upper limit. The average performance listed is calculated across all 10 benchmarks.}
    \vspace{-5pt}
    \label{tab:unified_trackers_dte_budget_ref}
    \centering
    \resizebox{\linewidth}{!}{
    \setlength{\tabcolsep}{2pt} 
    \begin{tabular}{l|ccc|cc|cc|cc|cc|cc|c}
        \toprule
        \multirow{2}{*}{Method} & 
        \multicolumn{3}{c|}{VOT-RGBD22} & 
        \multicolumn{2}{c|}{LasHeR} & 
        \multicolumn{2}{c|}{RGBT234} & 
        \multicolumn{2}{c|}{VisEvent} & 
        \multicolumn{2}{c|}{TNL2K} &
        \multicolumn{2}{c|}{OTB99} & 
        \multirow{2}{*}{\makecell{Avg\\Perf.(\%)}} \\
        \cmidrule(lr){2-4} \cmidrule(lr){5-6} \cmidrule(lr){7-8} \cmidrule(lr){9-10} \cmidrule(lr){11-12} \cmidrule(lr){13-14}
        & EAO & Acc & Rob       & AUC & P       & MSR & MPR         & AUC & P       & AUC & P       & AUC & P & \\
        \midrule
        \multicolumn{15}{c}{\textit{Upper Bound, 294 Tokens (100\%)}} \\
        \midrule
        \multirow{2}{*}{SUTrack\textsubscript{224}} & 75.5 & 82.5 & 91.3 & 59.9 & 74.8 & 70.0 & 92.1 & 63.3 & 80.7 & 67.8 & 73.8 & 68.4 & 91.1 & \multirow{2}{*}{100\%} \\
                                                    &100\% &100\% &100\% &100\% &100\% &100\% &100\% &100\% &100\% &100\% &100\% &100\% &100\% & \\ \midrule
        \multicolumn{15}{c}{\textit{Retain 71.4\% Tokens ($\approx$ 218 tokens, $\downarrow$ 25.6\%)}} \\ \midrule
        \multirow{2}{*}{SUTrack-CE\textsubscript{224}} & 75.8 & 82.8 & 91.4 & 59.9 & 74.7 & 70.0 & 92.5 & 61.8 & 79.2 & 66.0 & 71.5 & 70.5 & 92.6 & \multirow{2}{*}{\underline{99.5\%} ($\downarrow0.5\%$)} \\
                                                       &100.4\% &100.4\% &100.1\% &100.0\% &99.9\% &100.0\% &100.4\% &97.6\% &98.1\% &97.3\% &96.9\% &103.0\% &101.7\% & \\
        \cmidrule(lr){1-15}
        \multirow{2}{*}{SUTrack-ToMe\textsubscript{224}} & 76.0 & 82.5 & 91.9 & 60.0 & 74.6 & 69.0 & 92.0 & 62.3 & 79.6 & 66.2 & 71.8 & 70.4 & 91.3 & \multirow{2}{*}{99.4\%($\downarrow0.6\%$)} \\
                                                         &100.7\% &100.0\% &100.7\% &100.2\% &99.7\% &98.6\% &99.9\% &98.4\% &98.6\% &97.6\% &97.3\% &102.9\% &100.3\% & \\
        \cmidrule(lr){1-15}
        \multirow{2}{*}{SUTrack-EViT\textsubscript{224}} & 75.7 & 82.6 & 91.4 & 59.0 & 73.2 & 70.0 & 92.0 & 62.0 & 79.1 & 65.7 & 71.1 & 70.7 & 91.4 & \multirow{2}{*}{99.0\%($\downarrow1.0\%$)} \\
                                                         &100.3\% &100.1\% &100.1\% &98.5\% &97.9\% &100.0\% &99.9\% &97.9\% &98.0\% &96.9\% &96.4\% &103.4\% &100.4\% & \\
        \cmidrule(lr){1-15}
        \multirow{2}{*}{SUTrack-DyViT\textsubscript{224}} & 74.4 & 81.5 & 91.2 & 57.6 & 71.8 & 68.7 & 90.8 & 61.3 & 79.4 & 64.0 & 65.5 & 71.5 & 93.6 & \multirow{2}{*}{96.5\%($\downarrow3.5\%$)} \\
                                                          &98.5\% &98.8\% &99.9\% &96.2\% &96.0\% &98.1\% &98.6\% &96.8\% &98.4\% &94.4\% &88.9\% &104.5\% &102.8\% & \\
        \cmidrule(lr){1-15}
        \multirow{2}{*}{\textbf{UTPTrack\textsubscript{224}}} & 76.5 & 82.7 & 92.1 & 60.1 & 74.9 & 70.2 & 93.1 & 62.0 & 79.2 & 66.4 & 72.0 & 70.7 & 93.1 & \multirow{2}{*}{\textbf{99.8\%}($\downarrow0.2\%$)} \\
                                                     &101.3\% &100.4\% &100.9\% &100.3\% &100.1\% &100.3\% &101.1\% &97.9\% &98.1\% &97.9\% &97.6\% &103.3\% &102.2\% & \\
        \midrule
        \multicolumn{15}{c}{\textit{Retain 52.0\% Tokens ($\approx$ 153 tokens, $\downarrow$ 48.0\%)}} \\
                \midrule
        \multirow{2}{*}{SUTrack-CE\textsubscript{224}} & 76.1 & 82.8 & 91.8 & 58.4 & 72.7 & 70.4 & 93.0 & 61.5 & 78.6 & 66.1 & 71.7 & 70.5 & 92.6 & \multirow{2}{*}{\underline{99.2\%}($\downarrow0.8\%$)} \\
                                                       &100.8\% &100.4\% &100.5\% &97.5\% &97.2\% &100.6\% &101.0\% &97.2\% &97.4\% &97.5\% &97.2\% &103.0\% &101.7\% & \\
        \cmidrule(lr){1-15}
        \multirow{2}{*}{SUTrack-ToMe\textsubscript{224}} & 76.1 & 82.7 & 91.8 & 59.2 & 73.7 & 69.4 & 91.8 & 61.7 & 78.9 & 65.9 & 71.4 & 70.5 & 91.6 & \multirow{2}{*}{99.0\%($\downarrow0.5\%$)} \\
                                                         &100.8\% &100.2\% &100.5\% &98.8\% &98.5\% &99.1\% &99.7\% &97.5\% &97.8\% &97.1\% &96.9\% &103.1\% &100.6\% & \\
        \cmidrule(lr){1-15}
        \multirow{2}{*}{SUTrack-EViT\textsubscript{224}} & 75.2 & 82.1 & 91.4 & 57.8 & 71.9 & 69.9 & 92.3 & 62.3 & 79.8 & 65.1 & 70.3 & 70.5 & 91.7 & \multirow{2}{*}{98.5\%($\downarrow1.5\%$)} \\
                                                         &99.6\% &99.5\% &100.1\% &96.5\% &96.1\% &99.9\% &100.2\% &98.4\% &98.9\% &96.0\% &95.4\% &103.1\% &100.7\% & \\
        \cmidrule(lr){1-15}
        \multirow{2}{*}{SUTrack-DyViT\textsubscript{224}} & 63.0 & 74.3 & 84.8 & 50.6 & 63.4 & 63.3 & 86.3 & 49.6 & 67.2 & 54.3 & 54.4 & 65.3 & 86.0 & \multirow{2}{*}{78.8\%($\downarrow21.2\%$)} \\
                                                          &83.4\% &90.1\% &92.9\% &84.5\% &84.8\% &90.4\% &93.7\% &78.4\% &83.3\% &80.0\% &73.7\% &95.5\% &94.4\% & \\
        \cmidrule(lr){1-15}
        \multirow{2}{*}{\textbf{UTPTrack\textsubscript{224}}} & 75.7 & 82.8 & 91.4 & 59.6 & 74.4 & 70.7 & 93.6 & 61.7 & 78.8 & 65.9 & 71.5 & 70.1 & 91.2 & \multirow{2}{*}{\textbf{99.5\%}($\downarrow0.5\%$)} \\
                                                     &100.3\% &100.4\% &100.1\% &99.5\% &99.5\% &101.0\% &101.6\% &97.5\% &97.6\% &97.2\% &96.9\% &102.5\% &100.1\% & \\
        \midrule
        \multicolumn{15}{c}{\textit{Retain 35.4\% Tokens ($\approx$ 104 tokens, $\downarrow$ 64.6\%)}} \\
                \midrule
        \multirow{2}{*}{SUTrack-CE\textsubscript{224}} & 75.5 & 82.2 & 91.8 & 57.8 & 71.9 & 69.3 & 92.2 & 61.4 & 78.7 & 65.4 & 70.5 & 71.7 & 94.2 & \multirow{2}{*}{\underline{98.3\%}($\downarrow1.7\%$)} \\
                                                       &100.0\% &99.6\% &100.5\% &96.5\% &96.1\% &99.0\% &100.1\% &97.0\% &97.5\% &96.3\% &95.5\% &104.8\% &103.5\% & \\
        \cmidrule(lr){1-15}
        \multirow{2}{*}{SUTrack-ToMe\textsubscript{224}} & 70.9 & 78.6 & 90.2 & 53.8 & 66.6 & 63.6 & 87.9 & 58.0 & 77.4 & 61.7 & 63.6 & 67.3 & 87.0 & \multirow{2}{*}{92.5\%($\downarrow7.5\%$)} \\
                                                         &93.9\% &95.3\% &98.8\% &89.8\% &89.0\% &90.9\% &95.4\% &91.6\% &95.9\% &91.0\% &86.3\% &98.4\% &95.6\% & \\
        \cmidrule(lr){1-15}
        \multirow{2}{*}{SUTrack-EViT\textsubscript{224}} & 75.0 & 81.9 & 91.1 & 56.4 & 70.3 & 67.8 & 90.2 & 60.8 & 78.4 & 64.4 & 69.3 & 70.4 & 91.5 & \multirow{2}{*}{96.7\%($\downarrow3.3\%$)} \\
                                                         &99.3\% &99.3\% &99.8\% &94.2\% &94.0\% &96.9\% &97.9\% &96.1\% &97.1\% &95.0\% &94.0\% &102.9\% &100.5\% & \\
        \cmidrule(lr){1-15}
        \multirow{2}{*}{SUTrack-DyViT\textsubscript{224}} & 13.7 & 49.7 & 23.6 & 10.3 & 15.5 & 17.9 & 32.4 & 7.3 & 14.8 & 15.9 & 13.5 & 13.6 & 20.8 & \multirow{2}{*}{14.7\%($\downarrow85.3\%$)} \\
                                                          &18.1\% &60.2\% &25.8\% &17.2\% &20.7\% &25.6\% &35.2\% &11.5\% &18.3\% &23.4\% &18.3\% &19.9\% &22.9\% & \\
        \cmidrule(lr){1-15}
        \multirow{2}{*}{\textbf{UTPTrack\textsubscript{224}}} & 76.1 & 83.0 & 91.5 & 58.6 & 72.9 & 70.4 & 92.9 & 61.5 & 79.1 & 66.0 & 71.4 & 70.3 & 91.9 & \multirow{2}{*}{\textbf{99.3\%}($\downarrow0.7\%$)} \\
                                                     &100.8\% &100.6\% &100.2\% &97.8\% &97.5\% &100.6\% &100.9\% &97.2\% &98.0\% &97.3\% &96.9\% &102.7\% &100.9\% & \\
        \bottomrule
    \end{tabular}
    }
    \vspace{-5pt}
\end{table*}

\newpage
\section{Visualization}
\label{app:vis}

\begin{figure*}[t!]
    \centering
    \includegraphics[width=\linewidth]{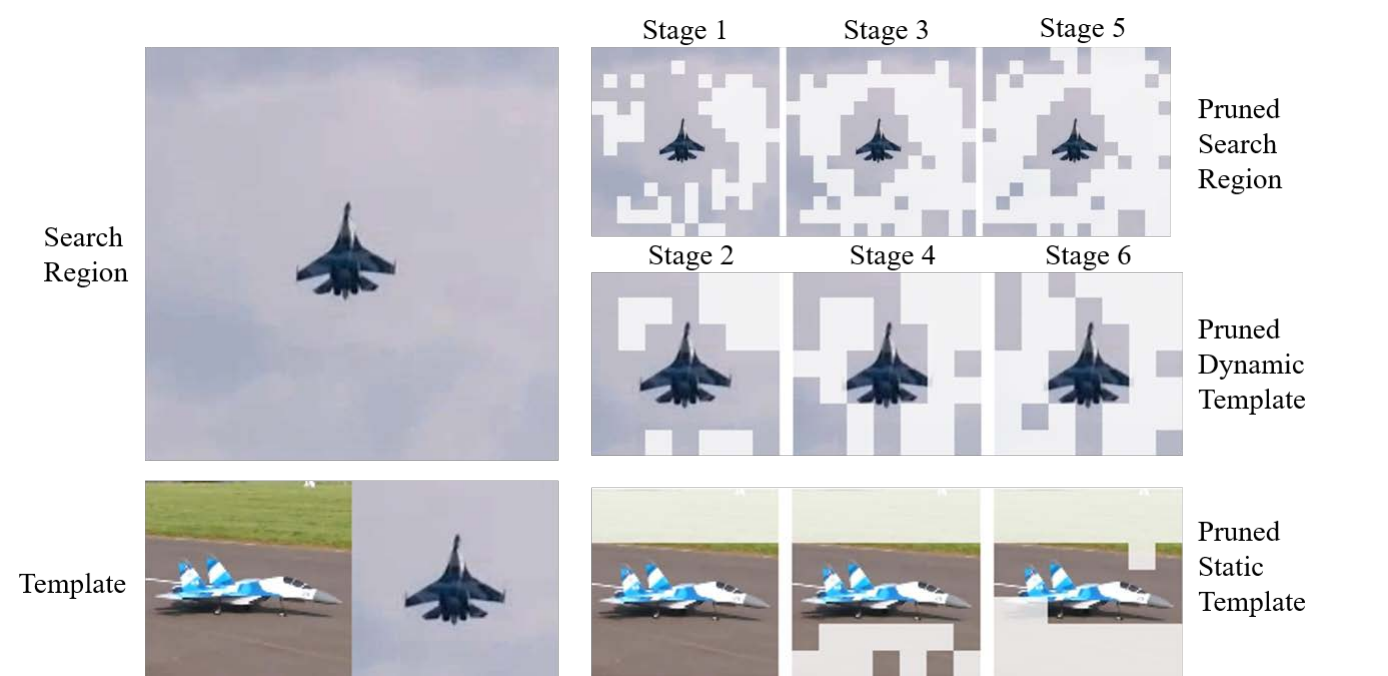} 
    \caption{Visualization of the UTPTrack Pruning Process for RGB-based Tracking.}
    \label{fig:rgb_ref}
\end{figure*}

\begin{figure*}[t!]
    \centering
    \includegraphics[width=\linewidth]{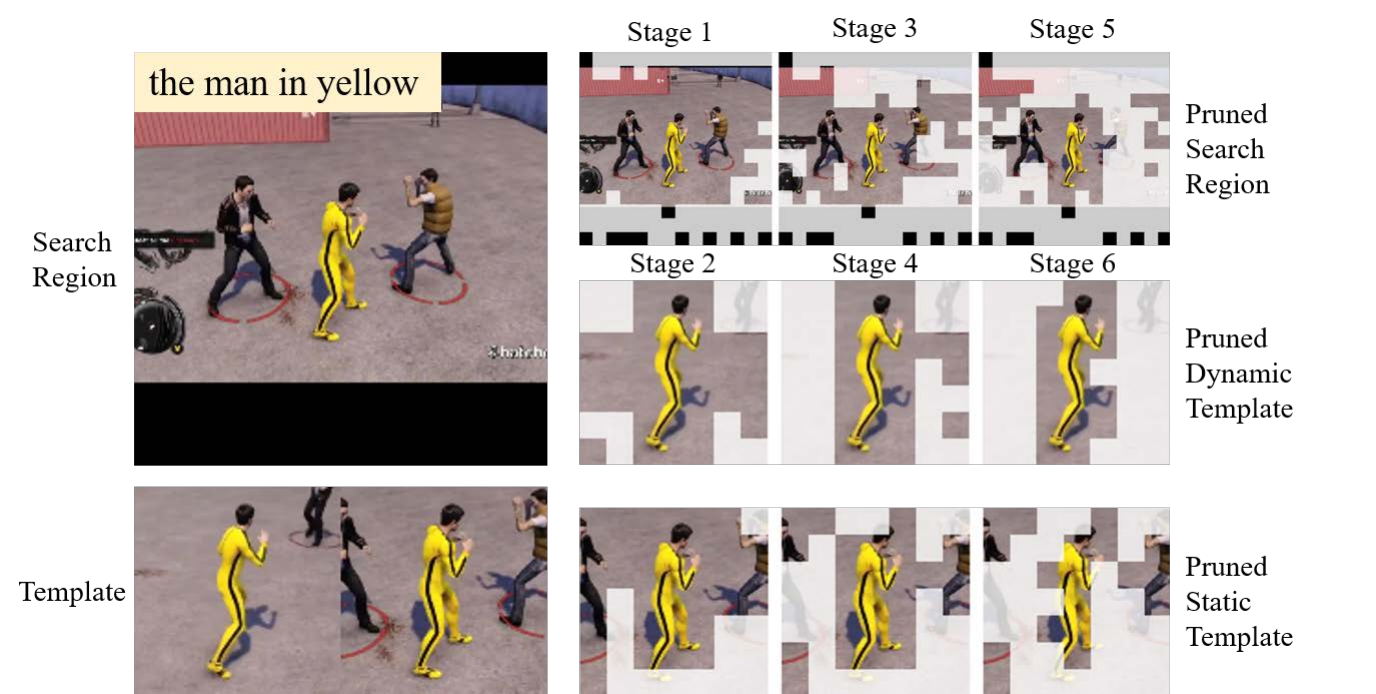} 
    \caption{Visualization of the UTPTrack Pruning Process for RGB-Language Tracking.}
    \label{fig:rgbl_ref}
\end{figure*}

\begin{figure*}[t!]
    \centering
    \includegraphics[width=\linewidth]{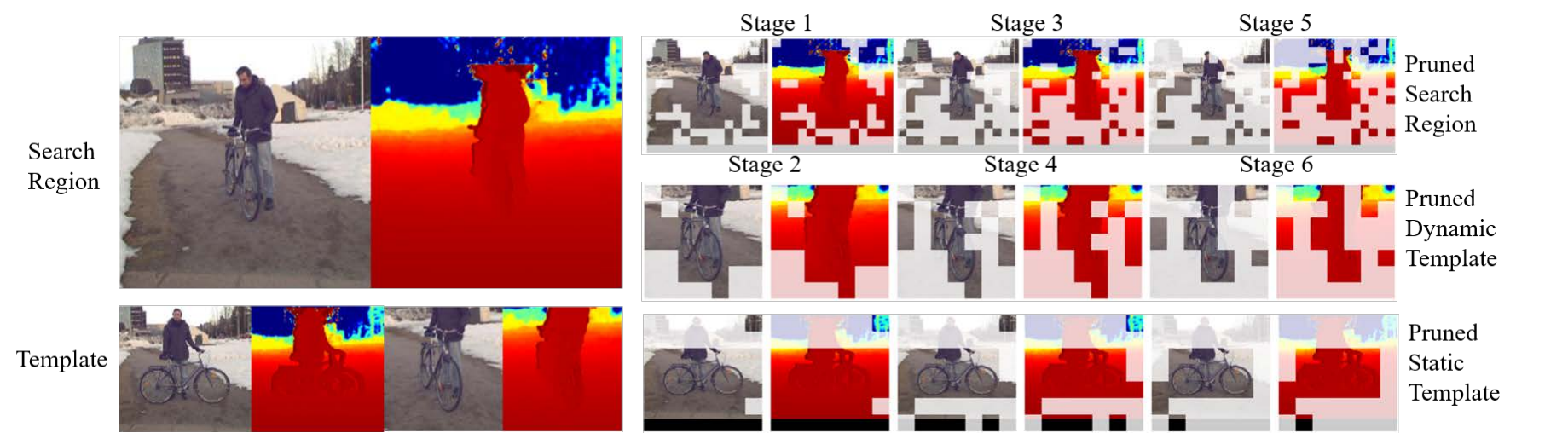} 
    \caption{Visualization of the UTPTrack Pruning Process for RGB-Depth Tracking.}
    \label{fig:rgbd_ref}
\end{figure*}

\begin{figure*}[t!]
    \centering
    \includegraphics[width=\linewidth]{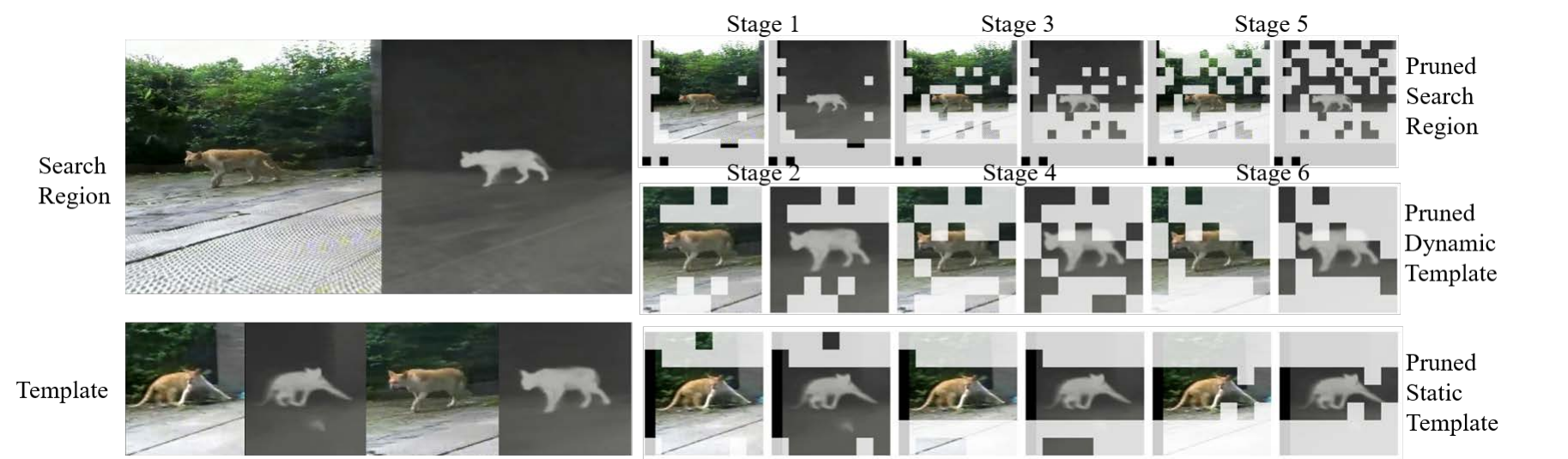} 
    \caption{Visualization of the UTPTrack Pruning Process for RGB-Thermal Tracking.}
    \label{fig:rgbt_ref}
\end{figure*}

\begin{figure*}[t!]
    \centering
    \includegraphics[width=\linewidth]{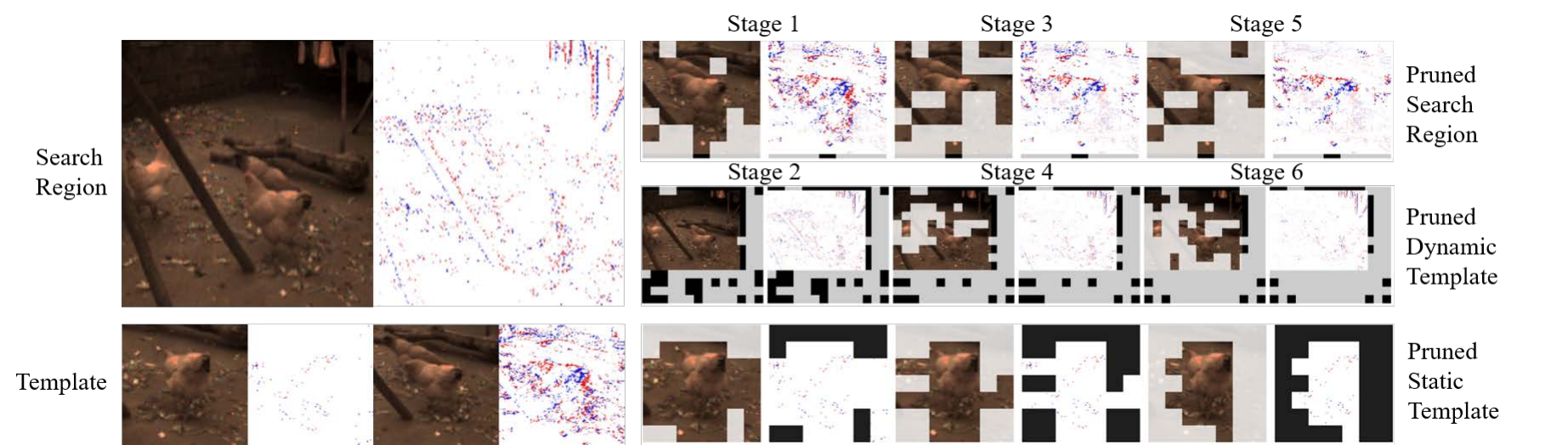} 
    \caption{Visualization of the UTPTrack Pruning Process for RGB-Event Tracking.}
    \label{fig:rgbe_ref}
\end{figure*}

We present visualizations of the UTPTrack pruning process for RGB-based, RGB-Lang, RGB-D, RGB-T, and RGB-E tracking, as shown in \cref{fig:rgb_ref,fig:rgbl_ref,fig:rgbd_ref,fig:rgbt_ref,fig:rgbe_ref}. The top-left shows the search region image, with the static template on the bottom-left and the dynamic template on the bottom-right. Masked areas indicate pruned tokens. The six stages illustrate a progressive schedule: stages 1, 3, and 5 progressively prune tokens from the search region, while stages 2, 4, and 6 further prune tokens from both the static and dynamic templates.

\end{document}